\documentclass[lettersize,journal]{IEEEtran} % this is for transactions in multimedia
\ifCLASSINFOpdf \else \fi

\usepackage{graphicx}
\usepackage{verbatim}
\usepackage[cmex10]{amsmath}
\usepackage{amssymb}
\usepackage{amsfonts}
\usepackage{bbm}
\usepackage{hyperref}
\usepackage{color}
\usepackage{float}
\usepackage{booktabs}

\usepackage{algorithmic}
\usepackage{algorithm2e}
\usepackage{setspace}

\usepackage{graphicx}
\usepackage{subcaption}

\usepackage{array}
\usepackage{multirow}
\usepackage{soul}
\usepackage{caption}
\usepackage{url}
\usepackage{graphicx}

\usepackage[noadjust]{cite}

\begin{document}

\title{S-HR-VQVAE: Sequential Hierarchical Residual Learning Vector Quantized Variational Autoencoder for Video Prediction}

\author{Mohammad Adiban$^\dagger$,
    Kalin Stefanov$^*$,
    Sabato Marco Siniscalchi$^{\dagger+}$~\IEEEmembership{Senior Member,~IEEE,}
    Giampiero Salvi$^\dagger$~\IEEEmembership{Senior Member,~IEEE}
\bigbreak

{$^\dagger$Norwegian University of Science and Technology\\
%Faculty of Information Technology and Electrical Engineering\\
$^*$Monash University\\%, Faculty of Information Technology\\
$^+$Università degli Studi di Palermo\\%, Dipartimento di Ingegneria\\
%, School of Electrical Engineering and Computer Science\\
E-mails: \{mohammad.adiban,marco.siniscalchi,giampiero.salvi\}@ntnu.no, kalin.stefanov@monash.edu}
} % end author

\maketitle

% in the abstract or keywords.
\begin{abstract}
We address the video prediction task by putting forth a novel model that combines (i) a novel hierarchical residual learning vector quantized variational autoencoder (HR-VQVAE), and (ii) a novel autoregressive spatiotemporal predictive model (AST-PM). We refer to this approach as a sequential hierarchical residual learning vector quantized variational autoencoder (S-HR-VQVAE). By leveraging the intrinsic capabilities of HR-VQVAE at modeling still images with a parsimonious representation, combined with the AST-PM's ability to handle spatiotemporal information, S-HR-VQVAE can better deal with major challenges in video prediction. These include learning spatiotemporal information, handling high dimensional data, combating blurry prediction, and implicit modeling of physical characteristics. Extensive experimental results on four challenging tasks, namely KTH Human Action, TrafficBJ, Human3.6M, and Kitti, demonstrate that our model compares favorably against state-of-the-art video prediction techniques both in quantitative and qualitative evaluations despite a much smaller model size. Finally, we boost S-HR-VQVAE by proposing a novel training method to jointly estimate the HR-VQVAE and AST-PM parameters.

\end{abstract}


\IEEEpeerreviewmaketitle

\section{Introduction}
\label{sec:introduction}
\textit{Video prediction} involves anticipating future video frames based on a sequence of preceding frames~\cite{vukotic2017one}. 
It is a challenging task, requiring algorithms to grasp complex spatiotemporal relationships within the video, 
%posing challenges in spatiotemporal modeling, 
at the same time as handling high dimensionality, addressing blurry predictions, and accounting for the physical characteristics of the scenes. 
\emph{Spatiotemporal modeling} aims to capture dependencies in video frame sequences, mirroring human perception of dynamic phenomena~\cite{Lu2017}.
This is a general problem in video modeling, but becomes especially challenging when we need to recursively and accurately predict video frames for long temporal spans.
Current state-of-the-art methods often struggle with long-term dependencies and complex motion patterns, leading to inaccuracies in the predicted frames.
\emph{High dimensionality} is inherent in video patterns, leading to the ``curse of dimensionality'' in function approximation and optimization~\cite{BELLMAN1961}.
Autoencoder-based methods attempt to reduce dimensionality, but may lose important fine-grained details necessary for accurate prediction.
\emph{Blurry predictions} stem from statistical models producing fuzzier outputs when predicting uncertain future events.
This is, therefore, a more challenging problem for video prediction than for any other video task.
Most methods use mean squared error (MSE) objective that tends to average over possible outcomes, resulting in blurred predictions.
The challenge of \emph{physical characteristics} pertains to object and scene attributes affecting prediction.
Proper modeling of these characteristics may potentially aid future frame predictions.
%%%%%%%%%%%%%%%%%%%%%%% added for the reviewer 1's suggestion to make more detailed introduction about challenges %%%%%%%%%%
Recent video prediction methods have made significant progress in tackling these challenges, yet they still face several limitations.
We will detail the state-of-the-art with respect to each of these challenges in Section~\ref{sec:related_work}.

This paper introduces a sequential hierarchical residual learning vector quantized variational autoencoder (S-HR-VQVAE), which is tailored for video prediction with the goal of tackling the above-mentioned challenges. 
To this end, S-HR-VQVAE implements a novel autoregressive spatiotemporal predictive model (AST-PM) to capture distributions of dependencies between latent representations across time and space. 
The latent representations are generated through our novel encoding scheme, termed hierarchical vector quantization variational autoencoder (HR-VQVAE) that we have recently used with success for still image reconstruction \cite{Adiban_2022_BMVC}.
Leveraging those two novel blocks, namely HR-VQVAE, and AST-PM, S-HR-VQVAE effectively tackles the video prediction task in three steps: 
In the first step, the input video frames are encoded to a continuous latent space and then mapped to discrete representations through HR-VQVAE, with each latent vector, in each layer in the model, assigned to a codeword in a codebook.
The key property of this model is the strict hierarchy imposed between codebooks belonging to different layers, producing extremely compact and efficient discrete representations.
In the second step, we predict future events in latent rather than image space.
To perform this prediction, we use spatiotemporal modeling (the proposed AST-PM), where the distribution of the discrete latent representations for a particular location in the current frame is conditioned on the representations for neighboring locations both in space and time.
In the third and final step, the predicted discrete representations are used by the HR-VQVAE decoder to generate the corresponding frame.
Normally, HR-VQVAE and AST-PM may be trained independently.
However, we also propose a novel joint training scheme to optimize HR-VQVAE and AST-PM together and show that this improves video prediction. 
We argue that the reason for the improved performance is that AST-PM and the decoder of HR-VQVAE are trained in such a way as to optimize both the predicted quantized latent representation for future frames as well as the reconstruction of future frames in image space.

Our contributions can be summarized as follows:
\begin{itemize}
\item  S-HR-VQVAE, a novel technique for video prediction, is proposed. This includes a hierarchical vector quantized encoding scheme and a spatiotemporal autoregressive model of the latent representations. This model allows the capture of different levels of abstraction in a sequence of video frames, thus resulting in a compact but effective representation of the task.
%\item A thorough and systematic experimental validation of the contributions of S-HR-VQVAE is presented while providing an extensive account and comparison with the state-of-the-art in video prediction on five video sequence prediction tasks, namely TrafficBJ~\cite{zhang2018predicting}, Human3.6M~\cite{ionescu2013human3}, Moving-MNIST~\cite{deng2012mnist}, KTH Human Action~\cite{schuldt2004recognizing} and Kitti~\cite{geiger2013vision}.
\item A novel loss function to jointly train the components of S-HR-VQVAE (HR-VQVAE and AST-PM) with further improvements of the prediction performance.
\item State-of-the-art results on several challenging video prediction tasks, namely KTH Human Action~\cite{schuldt2004recognizing}, TrafficBJ~\cite{zhang2018predicting}, Human3.6M~\cite{ionescu2013human3}  and Kitti~\cite{geiger2013vision}.%, as shown in Figure~\ref{fig:params}.
\end{itemize}

%\begin{figure}[ht]
%     \centering
%     \begin{subfigure}[ht]{0.24\textwidth}
%         \centering
%       \includegraphics[width=\textwidth]{figs/Moving-MNIST_MSE_scatter.pdf}
%         %\caption{MSE.}
%     \end{subfigure}
%     \hfill
%     \begin{subfigure}[ht]{0.24\textwidth}
%         \centering
%        \includegraphics[width=\textwidth]{figs/Moving-MNIST_SSIM_scatter.pdf}
         %\caption{PSNR.}
%     \end{subfigure}
%        \caption{Performance of S-HR-VQVAE on the MNIST moving dataset. The bright green arrow shows the direction of the model's performance with respect to the number of its parameters.
%}
%        \label{fig:params}
%\end{figure}

\section{Related Work}
\label{sec:related_work}

\subsection{Spatiotemporal Modeling}
Hu et al.~\cite{denton2017unsupervised} introduced disentangled representation net (DrNet) for spatial feature modeling in single video frames, neglecting temporal information. 
Motion-content network (McNet)~\cite{villegas2017decomposing} and mutual suppression
network (MsNet)~\cite{lee2018mutual} addressed motion and content separately, overlooking joint correlations. 
Convolutional long short-term memory (ConvLSTM)~\cite{shi2015convolutional} aimed at capturing both spatial and temporal correlations but struggled with long-term dependencies and scalability. 
To overcome ConvLSTM's limitations, Wang et al. proposed predictive recurrent neural network (PredRNN)~\cite{wang2017predrnn}, which, despite improvements, still faced challenges in modeling complex long-term dependencies. PredRNN++~\cite{wang2018predrnn++} and PredRNN-V2~\cite{wang2022predrnn} aimed to enhance PredRNN's performance by incorporating hierarchical recurrent structures. 
Eidetic 3D LSTM (E3D-LSTM)~\cite{wang2019eidetic} was introduced to jointly model spatial and temporal dynamics. Su et al.~\cite{su2020convolutional} improved efficiency using low-rank tensor factorization, while robust spatiotemporal LSTM (R-ST-LSTM)~\cite{saideni2022novel} and memory in memory (MIM)~\cite{wang2019memory} demonstrated performance improvements in long-term frame prediction tasks. 
The simple video prediction model (SimVP)~\cite{gao2022simvp} showed significant improvement over RNN-based models but struggled with encoding long-term dynamics, making accurate future prediction challenging.
%%paper asked by reviewer1.
Chang et al.~\cite{chang2022strpm} introduce hierarchical semantic separation in video prediction using a spatiotemporal encoding-decoding scheme and residual predictive memory called STRPM. This scheme separates spatial and temporal information with independent encoders, preserving distinct features and improving high-resolution video predictions. The STRPM refines the separation by focusing on inter-frame residuals for more accurate future predictions.
However, STRPM's reliance on residual inter-frame motion can oversimplify complex dynamics, and its implicit hierarchy may limit its ability to capture fine-grained spatiotemporal details compared to models with explicit multi-layered hierarchies.

To address the spatiotemporal challenge, S-HR-VQVAE leverages our proposed AST-PM module.
In this module, causal convolutions in time and spatiotemporal self-attention are used to model the spatiotemporal correlations on the quantized codes level.
Moreover, AST-PM operates on the latent discrete representations produced by the HR-VQVAE module instead of using pixels directly.

\subsection{High Dimensionality}
The above spatiotemporal methods rely on complex modeling, which hampers scalability, especially with the high dimensionality of video data. 
Hsieh et al. ~\cite{ranzato2014video} addressed this by dividing frames into patches and predicting their evolution over time using a recurrent convolutional neural network (rCNN) ~\cite{xu2019recurrent}. 
Jun-Ting et al.~\cite{hsieh2018learning} proposed the decompositional disentangled predictive autoencoder (DDPAE) framework, automatically breaking down high-dimensional videos into components with low-dimensional temporal dynamics. 
Xue et al. ~\cite{xue2016visual} proposed a variational autoencoder (VAE)~\cite{kingma2014stochastic} model to generate a distribution of next frame predictions. 
Oliu et al. ~\cite{oliu2018folded} utilized a folded recurrent neural network (fRNN) with a gated recurrent unit (GRU) for bidirectional information flow, enabling state sharing between the encoder and decoder. Variational 3D ConvLSTM (V-3D-ConvLSTM) ~\cite{razali2021log} combined variational encoder-decoder and 3D-ConvLSTM techniques. 
~\cite{ye2023video} developed thr video prediction
Transformer (VPTR), an attention-based encoder-decoder, to learn local spatiotemporal representations while simplifying the model.

Compared to the aforementioned methods, S-HR-VQVAE can effectively manage the high dimensionality of video data, leveraging the hierarchical structure inside the vector quantization module, which efficiently compresses each video frame, as demonstrated in the experimental section.

\subsection{Blurry Predictions}

As reported in~\cite{larsen2015autoencoding},  video prediction solutions quite often rely on RNNs, VAEs, and their variants (e.g., variational RNNs - VRNNs~\cite{castrejon2019improved}) resulting in blurry predictions.
Two main strategies have emerged to address this issue: (i) Latent variable methods that explicitly model underlying stochasticity and (ii) Adversarially-trained models that aim to produce more natural images.
In~\cite{denton2018stochastic}, the authors instead aimed to investigate stochastic models for video prediction using the VAE framework.
Given the recent advances in generative adversarial networks (GANs), researchers have also explored alternative techniques, such as VAE-GANs~\cite{larsen2015autoencoding, lee2018stochastic} for video frame prediction.
VAE-GANs allow capturing stochastic posterior distributions of videos while making it feasible to model the spatiotemporal joint distribution of pixels.
However, such methods often suffer from the problem of mode collapse and unrealistic predictions~\cite{lee2018stochastic, babaeizadeh2017stochastic}.

S-HR-VQVAE combats image blurring thanks to the temporal model leveraging the hierarchical codebook representation.
This allows for an increase in the quantization granularity without resulting in blurry images. In fact, despite the lossy nature of the compressed encoding, our experiments clearly demonstrate that the original video can be reconstructed with a high degree of fidelity through the latent representations.

\subsection{Physical Characteristics}
%Most of the discussed methods represent the information contained in the videos as a whole. However, videos are usually made of slowly varying contextual information, such as the background~\cite{barnich2009vibe} and a number of objects in the foreground that possibly vary more quickly following specific physical rules~\cite{primus2013segmentation}. For example, object velocity tends to be constant, and deformation in the human body can be modeled by joints and rods.

To leverage physical characteristics, some methods focus on pixel-level representations. 
For example, De Brabandere et al. ~\cite{jia2016dynamic} introduced the dynamic filter network (DFN), which learns local spatial transformations from flow information. 
Finn et al.~\cite{finn2016unsupervised} proposed convolutional dynamic neural advection (CDNA), a model that predicts object motion and pixel motion distributions from previous frames. In another approach ~\cite{liu2017video}, a system was developed to predict optical flows between future and past frames. 
Berg et al. ~\cite{jaderberg2015spatial} utilized backward content transformation via a 6-parameter affine model to learn future-to-past relationships. 
Villegas et al.~\cite{villegas2017decomposing}] employed LSTM to independently model pixel-level images for spatial layout and temporal dynamics, simplifying prediction tasks. 
Guen et al. introduced the Physical dynamics network (PhyDNet)~\cite{guen2020disentangling} to separate physical dynamics from other factors, yielding notable improvements. A motion-based modeling technique (MotionRNN)~\cite{wu2021motionrnn} decomposes motions into transient variations and trends, utilizing RNN-based models like ConvLSTM, PredRNN, and E3D-LSTM for prediction. 
Lee et al.~\cite{lee2021video} proposed a long-term motion context memory (LMC-memory) model for considering long-term motion context in future frame prediction. 
However, these methods primarily address physical characteristics, overlooking challenges like high dimensionality, blurry predictions, and spatiotemporal modeling in video prediction.

S-HR-VQVAE does not explicitly model physical characteristics.
Nonetheless, the modularity of the hierarchical vector quantization block allows S-HR-VQVAE to implicitly model physical characteristics.
In fact, latent representations are decomposed into a hierarchy of discrete codes, separating high-level global information (e.g., static background) from details (e.g., fine texture or small motions).
Since the latent representations are decomposed into different layers of hierarchical residual codes, the proposed AST-PM can exploit spatiotemporal dependencies that are different for different levels of detail.
For example, the background evolves slowly in time; whereas, the foreground object may move quickly.
Similarly, within the foreground object, some details, such as hands and arms, may exhibit different movement patterns compared to the body.
In sum, the combination of HR-VQVAE and AST-PM allows the modeling of physical characteristics, improving accuracy while reducing complexity.

\begin{figure*}
\centering
\includegraphics[width=0.9\linewidth]{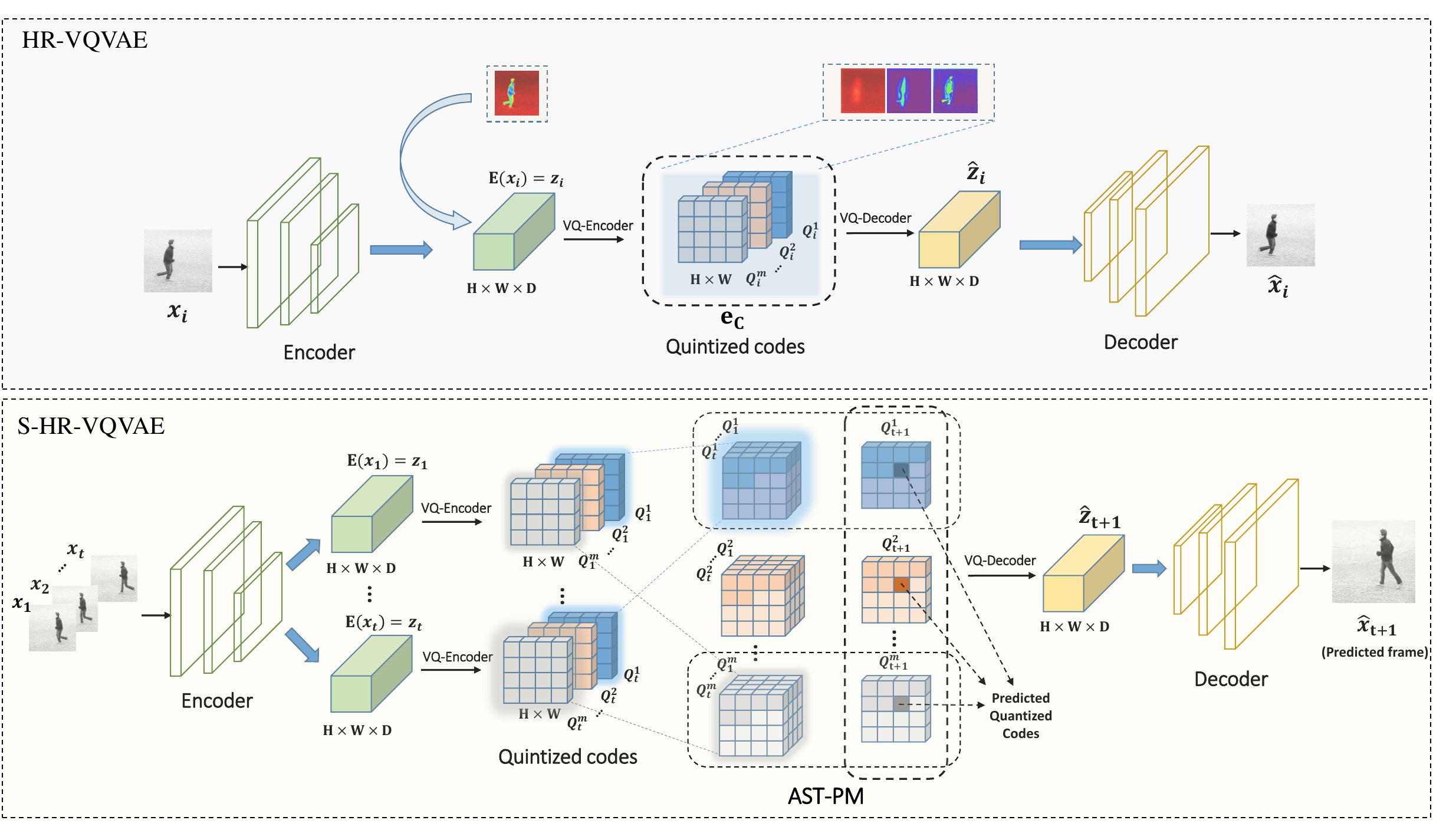}
\caption{Top: The HR-VQVAE module for hierarchical vector quantization, where each frame $i$ is encoded into $m$ hierarchical layers of quantized values $\left( Q_i^1, \dots, Q_i^m \right)$. Bottom: Illustration of the S-HR-VQVAE for video prediction, which combines HR-VQVAE with the AST-PM model. AST-PM predicts the indices of quantized values in both spatial and temporal dimensions, where each index at time $t+1$ is predicted by accessing only its preceding indices—those located above and to the left in a raster-scan spatial order and those before $t+1$ in the temporal domain.}
\label{fig:framework}
\end{figure*}

\section{Theoretical Background}
\label{sec:theoretical_background}

%\subsection{VQVAE}
%\label{sec:vqvae}
Variational autoencoders (VAEs) and vector quantized VAEs (VQVAEs) have been used for many applications for their inherent representation capabilities~\cite{doersch2016tutorial,van2017neural}.
Focusing on image processing applications, our primary focus, an input image is represented as a tensor $\mathbf{x} \in \mathbb{R}^\mathsmaller{{H_I\times W_I\times D_I}}$ of height $H_I$, width $W_I$ and $D_I$ color channels.
VQVAE first maps the input image $\mathbf{x}$ to a continuous latent vector $\mathbf{z}\in \mathbb{R}^\mathsmaller{{H\times W\times D}}$ through a non-linear encoder: $\mathbf{z} = E(\mathbf{x}) $.
Next, each element $\mathbf{z}_{hw} \in \mathbb{R}^\mathsmaller{D}$, with $h\in[1, H]$, and  $w\in[1, W]$, in the continuous latent vector $\mathbf{z}$ is quantized to the nearest codebook vector (i.e., a codeword) $\mathbf{e}_k\in \mathbb{R}^\mathsmaller{D},~k \in 1,...,m$ by
\begin{equation}
\label{eq:quantize_vqvae}
    \text{Quantize}(\mathbf{z}_{hw}) := \mathbf{e}_k \text{ where } k=\arg \min_{j}
    \|\mathbf{z}_{hw}-\mathbf{e}_j\|_2.
\end{equation}
The quantized vectors corresponding to each element $\mathbf{z}_{hw}$ are then recombined into the continuous representation $\mathbf{e} \in \mathbb{R}^\mathsmaller{{H\times W\times D}}$ to form the input of the decoder that reconstructs the input image using a transformation $\mathcal{D(\cdot)}$.
The loss function $\mathcal{L}(.)$ aims at minimizing the reconstruction error $\lVert\mathbf{x}-\mathcal{D}(\mathbf{e})\rVert_2$ whilst minimizing the quantization error $\lVert\mathbf{z}-\mathbf{e}\rVert_2$ as follows
\begin{equation}
\label{eq:loss_vqvae}
\mathcal{L}(\mathbf{x},\mathcal{D}(\mathbf{e})) = \lVert\mathbf{x}-\mathcal{D}(\mathbf{e})\rVert^2_2 + \lVert\text{sg}[\mathbf{z}] - \mathbf{e}\rVert^2_2 + \beta\lVert\text{sg}[\mathbf{e}] - \mathbf{z}\rVert^2_2,
\end{equation}
where $\text{sg}(.)$ is a stop-gradient operator cutting gradient flow during backpropagation, and $\beta$ is a hyperparameter governing the stability of encoder output latent vectors.

In \cite{razavi2019generating} a multi-layer version of VQVAE was proposed.
However, the representations at different levels in the architecture were not related hierarchically.

\section{Proposed Method}
\label{sec:method}

In \cite{Adiban_2022_BMVC}, we introduced a truly hierarchical version of VQVAE (HR-VQVAE) that is one of the building blocks of the video prediction method proposed in this work.
HR-VQVAE deals with limitations in techniques such as VQVAE, e.g., codebook collapse and non-locality in codewords' indices.
In HR-VQVAE, each layer captures residual information that is not properly modeled by the preceding layers, and the codebooks at different layers are constrained by a strict hierarchy.
%The aspects of HR-VQVAE that are relevant to the proposed method will be detailed in Section~\ref{sec:method}.
%Further details on this method can be found in \cite{Adiban_2022_BMVC}.

Fig.~\ref{fig:framework} shows the proposed framework.
Given $T$ input frames $(\mathbf{x}_1,\dots,\mathbf{x}_T)$ in a video, the goal is to predict the following $S$ frames $(\mathbf{x}_{T+1},\dots,\mathbf{x}_{T+S})$ in three steps.
%The approach follows three steps.
First, the input frames are encoded into a discrete latent representation using HR-VQVAE.
Next, a novel autoregressive spatiotemporal predictive model (AST-PM) is proposed to predict new discrete latent variables of future frames based on the latent variables for previous frames.
Finally, the HR-VQVAE decoder is used to generate the new frames from the latent variables obtained by AST-PM.
%Next, we will detail each step and describe two methods for training HR-VQVAE and AST-PM, either independently or jointly.
The proposed approach is referred as sequential hierarchical residual learning vector quantized variational autoencoder (S-HR-VQVAE).

\subsection{Step 1: Frame Encoding}
\label{sec:method_frame_encoding}
In the first step, each frame $\mathbf{x} \in \mathbb{R}^{H_I\times W_I\times D_I}$ is encoded using HR-VQVAE into a discrete latent representation.
HR-VQVAE first encodes the frame into a continuous vector $\mathbf{z} = E(\mathbf{x}) \in \mathbb{R}^{H\times W\times D}$.
These vectors are then iteratively quantized into $n$ hierarchical layers of discrete latent embeddings.
Assuming that the first layer has a single codebook of size $M$, the second layer has $M$ independent codebooks of size $M$ (for a total of $M^2$ codewords), and so on.
A generic layer $i$ has thereby $M^{i-1}$ codebooks of size $M$, for a total of $M^i$ codewords.
However, only one of those codebooks is used in each layer depending on which codewords were chosen in the previous layers.
In each layer $i$, the codebook is optimized to minimize the error between codewords $\mathbf{e}^i_k \in \mathbb{R}^\mathsmaller{D}$ and elements $\boldsymbol{\xi}^{i-1}_{hw} \in \mathbb{R}^\mathsmaller{D}$ of the residual error from the previous layer\footnote{For the first layer, $\boldsymbol {\xi}^0_{hw} \equiv \mathbf{z}_{hw}$.}
\begin{equation}
\label{eq:quantize_MLVAE}
    \text{Quantize}^i(\boldsymbol{\xi}^{i-1}_{hw}) := \mathbf{e}^i_k \text{ where } k=\arg \min_{j} \lVert\boldsymbol{\xi}^{i-1}_{hw}-\mathbf{e}^i_j\rVert_2,
\end{equation}
and $\mathbf{e}_k^i$ belongs to one of the possible codebooks $C_{i}(t)$ for layer $i$.
Which codebook is used is determined by the codeword $\mathbf{e}_t^{i-1}$ selected at the previous layer.
Within each layer, the codewords $\mathbf{e}^i_k$, for each element $\boldsymbol{\xi}^{i-1}_{hw}$ of the residual, are combined to form the tensor $\mathbf{e}^i \in \mathbb{R}^\mathsmaller{{H\times W\times D}}$.
Across the different layers, the tensors $\mathbf{e}^i$ are then summed to form the ``combined'' discrete representation $\mathbf{e}_C$.
When HR-VQVAE is used to reconstruct single images, $\mathbf{e}_C$ is fed into the decoder to reconstruct the image as $\mathbf{\hat{x}} = \mathcal{D}(\mathbf{e}_C)$, and the corresponding objective function is used to train the system

% this is the two column version
\begin{multline}
\label{eq:loss_mlvae}
\mathcal{L}({\mathbf{x}},\mathcal{D}({\mathbf{e}_C})) = \lVert{\mathbf{x}}-\mathcal{D}({\mathbf{e}_C})\rVert^2_2 + \lVert\text{sg}[\boldsymbol{\xi}^0] -  {\mathbf{e}_C}\rVert^2_2 \\
    +\beta_0\lVert\text{sg}[{\mathbf{e}_C}] - \boldsymbol{\xi}^0\rVert^2_2 +\sum_{i=1}^{n}\mathcal{L}(\boldsymbol{\xi}^{i-1}, \mathbf{e}^i),
\end{multline}
with
\begin{equation}
    \label{eq:loss_layer}
    \mathcal{L}(\boldsymbol{\xi}^{i-1}, \mathbf{e}^i) = \lVert\text{sg}[\boldsymbol{\xi} ^{i-1}] - \mathbf{e}^i\rVert^2_2 
    +\beta_i \lVert\text{sg}[\mathbf{e}^i] - \boldsymbol{\xi}^{i-1}\rVert^2_2.
\end{equation}
The $\beta_i$ are hyperparameters that control the reluctance to change the code corresponding to the encoder output.
The main goal of Eqs.~\ref{eq:loss_mlvae} and \ref{eq:loss_layer} is to make a hierarchical mapping of input data in which each layer of quantization extracts residual information from its bottom layers.

In the proposed S-HR-VQVAE, we do not reconstruct images directly.
The indices to the codewords $\mathbf{e}^i$ are, instead, used as latent representations for each input frame in the video and each layer in the system and are input to the video prediction steps described below.
We call these indices for layer $i$, $Q^i \in [1, M]^{H\times W}$, with $M$ the size of each codebook.
The difference is clarified in Figure \ref{fig:framework}, where the image reconstruction case is represented in the top panel, and the video prediction is depicted in the bottom panel.

\subsection{Step 2: Spatiotemporal Latent Representation Prediction}
In the second step, the indices $(Q^i_1, \dots, Q^i_T)$ of the codewords $(\mathbf{e}^i_1, \dots, \mathbf{e}^i_T)$, obtained from each layer $i$ of HR-VQVAE from the input frames $(\mathbf{x}_1, \dots, \mathbf{x}_T)$, are used to predict the indices $(Q^i_{T+1}, \dots, Q^i_{T+S})$ of the codewords $(\mathbf{e}^i_{T+1},\dots, \mathbf{e}^i_{T+S})$ for $S$ future frames, with the goal of predicting the $S$ next future frames $(\mathbf{x}_{T+1},\dots, \mathbf{x}_{T+S})$.

To this end, we propose a probabilistic autoregressive spatiotemporal predictive model (AST-PM).
AST-PM takes discrete indices of the latent representations as input and predicts future indices.
% \textcolor{red}{Combined with the hierarchical nature of HR-VQVAE, this considerably simplifies the spatial and temporal prediction issue, and our model focuses on essential aspects of the frames in space and time.
% Therefore, the model predicts the future codeword indices $\hat{Q}_{t>T}$ using the codeword indices at previous times ($Q^i_1, \dots, Q^i_T$).
% To explain the proposed probabilistic model, we first order the elements of $Q^i_t \in [1,M]^{H \times W}$, from left to right and from top to bottom using a linear index $v_k \in [1,HW]$.
% Then, we use the notation $v_{j<k}$ to refer to any element of $Q^i_t$ to the left or the top of $v_k$.
% Given the above notation, the probabilistic model can be written as
% \begin{equation}
%     \label{eq:rnn}
%     p(\hat{Q}_{t+1}^i(v_k)) = \prod^{H\times W}_{j=1} p(\hat{Q}_{t+1}^i(v_{j<k}) | Q^i_1(v_{j<k}),\dots,Q^i_t(v_{j<k})),
% \end{equation}
% where $\hat{Q}^i_t$ represents the predicted quntized discrete codes of layer $i$ obtained from the $t^{th}$ frame.
% The above behavior is obtained by using convolutional masks to limit the information that is used during prediction.
% The convolutional masks constrain the convolutions to retrieve only spatial information from the left and above each pixel.
% For the temporal dimension, convolutions were restricted to previous time steps by masking out present and future timesteps.
% This strategy is implemented using multi-head attention layers analogous to~\cite{razavi2019generating}.
% However, the attention is applied to 3D voxels here.}
Because HR-VQVAE is hierarchical, it greatly simplifies predicting spatial and temporal information, which allows our model to focus on the most important parts of the frames in both space and time.
Accordingly, the model predicts future codeword indices $\hat{Q}{t>T}$ using the codeword indices from previous times ($Q^i_1, \dots, Q^i_T$). To explain our probabilistic model, we first arrange the elements of $Q^i_t \in [1,M]^{H \times W}$ from left to right and top to bottom using a linear index $v_k \in [1, HW]$. We use the notation $v{j<k}$ to refer to any element of $Q^i_t$ that is to the left or above $v_k$.
Given this notation, the probabilistic model can be written as:
\begin{equation}
    \label{eq:rnn}
    p(\hat{Q}_{t+1}^i(v_k)) = \prod^{H\times W}_{j=1} p(\hat{Q}_{t+1}^i(v_{j<k}) | Q^i_1(v_{j<k}),\dots,Q^i_t(v_{j<k})),
\end{equation}
where $\hat{Q}^i_t$ represents the predicted quantized discrete codes of layer $i$ obtained from the $t^\text{th}$ frame.
This behavior is achieved by using convolutional masks to limit the information used during prediction. The convolutional masks restrict the convolutions to retrieve only spatial information from the left and above each index. For the temporal dimension, convolutions are limited to previous time steps by masking out present and future timesteps. This strategy is implemented using multi-head attention layers similar to those in~\cite{razavi2019generating}. However, in our case, the attention is applied to 3D voxels.
The loss function of AST-PM is as follows
\begin{multline}
\label{eq:pixelcnn_loss}
    \mathcal{L}_{p}(p(Q^i_{t>T}), p(\hat{Q}^i_{t>T})) = \\ - \frac{1}{H \times W} \sum_{j=1}^{H\times W}\sum_{m=1}^{M} p(Q^i_{t>T}) [j,m]  * \log  p(\hat{Q}^i_{t>T})[j,m],
\end{multline}

\subsection{Step 3: Frame Generation}
Once the quantization indices $\hat{Q}^i_t$ for each layer $i$ and each time step $t \in [T+1, T+S]$ have been estimated by the AST-PM, the corresponding quantized representation $\mathbf{\hat{z}}_t \in \mathbb{R}^{H\times W\times D}$ can be computed by codebook access ${\mathbf{e}_{C(t)}} = \sum\limits_{i=1}^{m} e^i_t$ (see Section~\ref{sec:method_frame_encoding}).

Finally, the predicted quantized codes are decoded to sequences of frames using the HR-VQVAE decoder $\mathcal{D}(.)$
\begin{equation}
    \label{eq:decode}
    (\hat{\mathbf{x}}_{T+1}, ..., \hat{\mathbf{x}}_{T+S}) = (\mathbf{\mathcal{D}(\hat{z}}_{T+1}),..., \mathcal{D}(\mathbf{\hat{z}}_{T+S})),
\end{equation}
where the and $\hat{\mathbf{z}}_{t>T}$ and $\hat{\mathbf{x}}_{t>T}$ represent the predicted latent representations and frames, respectively.

\subsection{Disjoint and Joint Training}
HR-VQVAE and AST-PM in the combined model described above can be trained independently.
In this case, we first train HR-VQVAE according to Eq.~\ref{eq:loss_mlvae} to predict each frame $\mathbf{x}_i$ in the video independently of the others.
We obtain a sequence of latent representations $(Q_1^i, \dots, Q_T^i, Q_{T+1}^i, \dots, Q_{T+S}^i)$ for each layer in HR-VQVAE and for the complete sequence of frames.
The AST-PM can the been trained to predict the sequence $(Q_{T+1}^i, \dots, Q_{T+S}^i)$ given the input sequence $(Q_1^i, \dots, Q_T^i)$, by optimizing Eq.~\ref{eq:pixelcnn_loss}.
In the test phase, we use the predictions of AST-PM in combination with the HR-VQVAE decoder to predict unseen video frames, making sure that the combined model only has access to $(\mathbf{x}_1, \dots, \mathbf{x}_T)$ when predicting $(\mathbf{x}_{T+1}, \dots, \mathbf{x}_{T+S})$.

Following this training procedure, the decoder in HR-VQVAE is exclusively optimized to deal with the uncertainty introduced by the encoder of HR-VQVAE, which means that the decoder is optimized solely for reconstructing the original input frame.
However, when we reconstruct the frames $(\mathbf{x}_{T+1}, \dots, \mathbf{x}_{T+S})$, we also need to deal with the uncertainty introduced by the AST-PM predictions. In fact, the AST-PM uncertainty refers to the mismatch between the predicted latent spaces of future frames and the actual latent space of future frames.
This indicates that the HR-VQVAE decoder block and the AST-PM block operate independently, without being aware of the uncertainties introduced by the other block.
In an attempt to address this issue, we propose to optimize the AST-PM and the HR-VQVAE decoder jointly.
Therefore, we proposed a joint training in Eq.~\ref{eq:joint_pixelCNN}
which includes two distinct objectives: the loss for the HR-VQVAE decoder, represented by the first term of Eq.~\ref{eq:loss_mlvae}, and the AST-PM loss in Eq.~\ref{eq:pixelcnn_loss}. 
The corresponding multi-objective loss is:

\begin{equation}
\label{eq:joint_pixelCNN}
        \mathcal{L}_{\text{joint}} = \mathcal{L}_{p} + \lambda \lVert{\mathbf{x}_t}-\mathcal{D}({\mathbf{e}_{C(t)}})\rVert^2_2,
\end{equation}
where $\lambda$ is a hyperparameter that controls the effect of the reconstruction loss on the joint training.
In this case, during training, HR-VQVAE only produces the latent representations $(Q_1^i, \dots, Q_T^i)$ for the input frames $(\mathbf{x}_1, \dots, \mathbf{x}_T)$.
The latent representations $(Q_{T+1}^i, \dots, Q_{T+S}^i)$ for the frames $(\mathbf{x}_{T+1}, \dots, \mathbf{x}_{T+S})$ are predicted by AST-PM and then used to train the HR-VQVAE decoder.

\section{Experiments}
\label{sec:experiments}

\subsection{Datasets}
We conducted experiments using four different challenging datasets.
Table~\ref{tab:dataset} presents a summary of corresponding statistics, including the number of training samples (\#Train), the number of test samples (\#Test), image resolution represented as $(H, W, C)$, input sequence length indicated as $T$, and predicted sequence length referred to as $\hat{T}$.

The \textbf{KTH Human Action} dataset~\cite{schuldt2004recognizing} is a moving image dataset with a resolution of $160\times120$ pixels that contains six types of human actions, including walking, jogging, running, boxing, hand waving, and hand clapping.
The dataset comprises 25 human subjects performing actions in four different scenarios.
For our experiments, we followed~\cite{wang2022predrnn}, resized the video frames down to $128\times128$, and split the dataset into two subsets: (i) a training set, consisting of the first 16 subjects, and (ii) a test set, containing the remaining subjects.

The \textbf{TrafficBJ} is a collection of taxicab GPS data and meteorological data recorded in Beijing~\cite{zhang2018predicting}.
Each frame in TrafficBJ has $32 \times 32$ pixels, including the traffic flow entering and leaving the same district.
We normalized the data to $[0, 1]$ and follow the experimental settings as~\cite{zhang2017deep}.

The \textbf{Human3.6M} dataset~\cite{ionescu2013human3} consists of 3.6 million samples capturing diverse human activities.
Similar to previous papers~\cite{gao2022simvp, guen2020disentangling, wang2019memory}, we focus on the ``walking'' scenario.

The \textbf{Kitti} dataset~\cite{geiger2013vision} was created through real traffic scenario collections by specially equipped vehicles, a joint effort by Germany's Karlsruhe Institute of Technology and the Toyota Institute of Technology in the United States.
We employ Kitti using three scenarios: road, city, and residential, resulting in 57 videos for a training set and 4 for a test set.
 %It should be noted that frames were center-cropped and downsampled to $128\times128$ pixels.

Finally, it is important to note that 5\% of the training set was reserved as a validation set, which was used specifically for fine-tuning the hyperparameters.

\begin{table}[t]
\scriptsize
\centering
\caption{Dataset statistics. \#Train and \#Test indicate the number of samples for the training and test set, respectively. Each input sequence consists of $T$, and the output sequence consists of $\hat{T}$ frames with shape (H, W, C).}
\label{tab:dataset}
\begin{tabular}{c|c|c|c|c|c}
\toprule[0.5mm]
\textbf{Dataset} & \textbf{\#Train} & \textbf{\#Test} & \textbf{(H, W, C)} & $\pmb{T}$ & $\pmb{\hat{T}}$\\
\midrule[0.25mm]
TrafficBJ~\cite{zhang2018predicting} & 19,627 & 1,334 & (32, 32, 2) & 4 & 4\\
%M-MNIST & 10,000 & 10,000 & (64, 64, 1) & 10 & 10\\
KTH~\cite{schuldt2004recognizing} & 5,200 & 3,167 & (128,128, 3) & 10 & 20\\
Human3.6M~\cite{ionescu2013human3} & 2,624 & 1,135 & (128, 128, 3) & 4 & 4\\
Kitti~\cite{geiger2013vision} & 40,783 & 1,963 & (128, 128, 3) & 4, 5 &5 \\
\bottomrule[0.5mm]
\end{tabular}
\end{table}

\subsection{Experimental Setup}
Table~\ref{tbl:compression} lists some details about S-HR-VQVAE architecture for tackling the datasets.
\emph{Input size} refers to the initial resolution of the video frames.
\emph{Latent size} corresponds to the continuous latent representation in HR-VQVAE.
\emph{Quantized latent size} to the quantized representation in the model.
We also provide additional information for the bit rate, number of hierarchy layers, codebook size, and number of codewords.

The proposed S-HR-VQVAE was trained on sequences consisting of 10 consecutive frames to predict 20 future frames for KTH Human Action, 
and also trained on 4 consecutive frames to predict 4 future frames for both TrafficBJ and Human3.6M datasets, which is a common practice for the tasks.
In addition, for the Kitti dataset, we focused on two specific settings: (i) 4 input frames and 5 predicted frames and (ii) 5 input frames and 5 predicted frames.
In all experiments, the model is trained using the Adam optimizer~\cite{kingma2014adam}, and the learning rate is set to  0.0003 for both HR-VQVAE encoder-decoder and AST-PM.
Besides, $\lambda$ in Eq.~\ref{eq:joint_pixelCNN} is set to 0.11.

\begin{table}[t]
\setlength{\tabcolsep}{2.5pt}
\scriptsize
\centering
\caption{Configuration details for S-HR-VQVAE.}
\label{tbl:compression}
\begin{tabular}{lccccccccc}
\toprule[0.5mm]
& \multicolumn{3}{c}{\textbf{KTH \& Human3.6M \& Kitti}} & \multicolumn{3}{c}{\textbf{TrafficBJ}} \\
\cmidrule[0.25mm](lr){2-4}\cmidrule[0.25mm](lr){5-7}
Input size & \multicolumn{3}{c}{$128\times128$} & \multicolumn{3}{c}{$32\times32$} \\
Bit rate & \multicolumn{3}{c}{8} & \multicolumn{3}{c}{8} \\
Latent size & \multicolumn{3}{c}{$32\times32\times8$} & \multicolumn{3}{c}{$16\times16\times4$} \\
Quantized size & \multicolumn{3}{c}{$32\times32$} & \multicolumn{3}{c}{$16\times16$} \\
\cmidrule[0.25mm](lr){2-4}\cmidrule[0.25mm](lr){5-7}
\#Layers & 1 & 3 & 9 & 1 & 3 & 6 \\
Codebook size & 512 & 8 & 2 & 64 & 4 & 2 \\
\#Codewords & 512 & \{8, 64, 512\} & \{2, 4,..., 512\} & 64 & \{4, 16, 64\} & \{2, 4,..., 64\} \\
%Compression ratio & $\approx$42.66 & $\approx$42.66& $\approx$42.66 & $\approx$21.33 &$\approx$21.33 &$\approx$21.33& $\approx$21.33 &$\approx$21.33 &$\approx$21.33 \\
\bottomrule[0.5mm]
\end{tabular}
\end{table}

\subsection{Metrics}
\label{sec:metrics}
We report results adopting metrics that are commonly used in the literature, namely: 
peak signal-to-noise ratio (PSNR)~\cite{winkler2008evolution}, structural similarity index measure (SSIM)~\cite{sitzmann2019scene}, learned perceptual image patch similarity (LPIPS)~\cite{zhang2018unreasonable}, frechet video distance (FVD)~\cite{unterthiner2019fvd}, mean square error (MSE), and mean absolute error (MAE).
PSNR, SSIM, LPIPS, MSE, and MAE are all image quality metrics but differ in their characteristics.
PSNR focuses on signal-to-noise ratio, SSIM considers structural similarity, MSE and MAE measure pixel-wise differences, and LPIPS aims to capture perceptual similarity based on deep neural networks. 
FVD, on the other hand, is a comprehensive video quality metric employed to assess the quality of generated videos. This evaluation is achieved by quantifying the feature distribution gap between real and generated videos, which effectively captures both temporal inconsistencies and motion-related artifacts. Furthermore, FVD evaluates both the temporal coherence of video content and the quality of individual frames, offering a holistic perspective on video realism and overall coherence.
All those metrics, however, have limitations.
For example, PSNR, MSE, and MAE have been shown to have poor correlation with human perception~\cite{mrak2004reliability, wang2004image} and may not take into account higher-level semantic information, such as in action modeling.
SSIM and LPIPS are more effective in capturing perceptual differences, but they may not be sensitive to all types of visual information: they may not be as effective at capturing differences in color or texture as at capturing differences in luminance and contrast~\cite{fei2012perceptual}.
fFVD tends to prioritize a video's spatial elements and may overlook the natural flow of its temporal dynamics~\cite{kimstream}.
Therefore, several metrics must be considered to better capture different aspects of the video prediction task and obtain a more comprehensive assessment of the methods' performance.

We report results according to all those metrics and include all available results for the related methods.
Because of the limitations of these metrics, we also provide a qualitative assessment to verify whether the metrics have missed some important aspects of the video prediction task.

\section{Results}
\label{sec:results}
Results of the quantitative evaluation of the proposed method followed by a qualitative assessment are now presented.
To better appreciate the effectiveness of the proposed technique, we have performed a systematic review of reported quantitative results of recent, state-of-the-art solutions.% on the four datasets.

The qualitative analysis is performed by observing the behavior of the proposed method on several video sequences, which is a common practice in the research field.
However, while reviewing the literature, we noticed that different methods use different video sequences to visually demonstrate the quality of their approaches; furthermore, the source code is not available for all methods in the literature, which implies that different systems can not be compared on the same set of predefined video sequences.
To overcome that issue, we first selected video sequences common among different techniques in the literature.
Then, we evaluated our S-HR-VQVAE on those selected examples and grouped the results accordingly.
To the best of our knowledge, this is the first time that such a systematic comparison has been carried out.

We also provide results for other aspects of the proposed S-HR-VQVAE, including reconstruction capability in (i) blur mitigation, (ii) noise removal, and (iii) compression.

\begin{table}[t]
\footnotesize 
\setlength{\tabcolsep}{2.5pt}
\centering
\caption{Results on KTH Human Action dataset. S-HR-VQVAE with 3 layers was used with disjoint and joint training.}
\label{tbl:results_comparision_KTH_MNIST}

\resizebox{\columnwidth}{!}{%
\begin{tabular}{lccccc}
\toprule[0.5mm]
& \multicolumn{5}{c}{KTH Human Action (10 $\rightarrow$ 20)}  \\ 
\cmidrule[0.25mm](lr){2-6}
\textbf{Method} & \textbf{PSNR}$\uparrow$ & \textbf{SSIM}$\uparrow$ & \textbf{LPIPS}$\downarrow$ & \textbf{\#Params} & \textbf{FLOPs}\\
\midrule[0.25mm]
ConvLSTM (2015)~\cite{shi2015convolutional} & 23.01 & 0.704 & 0.156 & 16.60M &  1,468G\\
DFN (2016)~\cite{jia2016dynamic} & 27.26 & 0.794 & $\times$ & $\times$ & $\times$\\
CDNA (2016)~\cite{finn2016unsupervised} & 23.75 & 0.752 & $\times$ & $\times$ & $\times$\\
DrNet(2017)~\cite{denton2017unsupervised} & 25.56 & 0.764 & $\times$ & 23.30M & $\times$\\
PredRNN (2017)~\cite{wang2017predrnn} & 27.55 & 0.839 & 0.167 & 23.85M & 2,800G\\
McNet (2018)~\cite{villegas2017decomposing} & 25.95 & 0.804 & $\times$ & 3.50M & $\times$\\
MsNet (2018)~\cite{lee2018mutual} & 27.08 & 0.876 & $\times$ & 3.20M & $\times$\\
fRNN (2018)~\cite{oliu2018folded} & 26.12 & 0.771 & $\times$ & $\times$ & $\times$\\
PredRNN++ (2018)~\cite{wang2018predrnn++} & 28.62 & 0.888 & 0.229 & 15.40M & 4,162G\\
E3D-LSTM (2019)~\cite{wang2019eidetic} & 27.92 & 0.893 & $\times$ & 41.94M & 214.0G\\
MIM (2019)~\cite{wang2019memory} & 27.78 & 0.902  & 0.188  & 37.37M & 1,099G\\
Conv-TT-LSTM (2020)~\cite{su2020convolutional} & 28.36 & 0.907 & 0.133 & 39.8M & $\times$\\
PhyDNet (2020)~\cite{guen2020disentangling} & 28.69 & $\times$ & 0.188 & 3.10M  & \textbf{93.6G}\\
Jin et al. (2020)~\cite{jin2020exploring} & 29.85 & 0.893 & 0.118 & $\times$ & $\times$\\
LMC-Memory (2021)~\cite{lee2021video} & 28.61 & 0.894 & 0.133 & $\times$ & $\times$\\
V-3D-ConvLSTM (2021)~\cite{razali2021log} & 28.31 & 0.866 & $\times$ & 12.90M & $\times$\\
R-ST-ConvLSTM (2022)~\cite{saideni2022novel} & 28.99 & 0.854 & $\times$ & $\times$ & $\times$\\
SimVP (2022)~\cite{gao2022simvp} & \textbf{33.72} & 0.905 & $\times$ & 22.30M & 125.6G\\
PredRNN-V2 (2023)~\cite{wang2022predrnn} & 28.37 & 0.838 & 0.139 & 23.86M & 2,815G\\
VPTR (2023)~\cite{ye2023video} & 26.96 & 0.879 & 0.076 & 162.48M & $\times$\\
NPVP (2023)~\cite{ye2023unified}$^*$ & 27.66 & 0.909 & \textbf{0.066} & $\times$ & $\times$\\
\midrule[0.25mm]
S-HR-VQVAE-disjoint (ours) & 28.43 & 0.863 & 0.130 & \textbf{1.14M} & 94.1G\\
S-HR-VQVAE-joint (ours) & 28.49 & \textbf{0.910} & 0.093 & \textbf{1.14M} & 95.8G\\
\bottomrule[0.5mm]
\multicolumn{5}{l}{$^*$ NPVP resized KTH samples to $64\times64$ instead of standard $128\times128$.} \\
\multicolumn{5}{l}{($\uparrow$) means higher is better and ($\downarrow$) means lower is better.}
\end{tabular}%
}
\end{table}

\begin{table*}
\scriptsize 
\setlength{\tabcolsep}{10pt}
\centering
\caption {Results on TrafficBJ and Human 3.6M. S-HR-VQVAE with 3 layers was used with disjoint and joint training.}
\label{tbl:results_comparision_traffic_human}
\begin{tabular}{lccccccccc}
\toprule[0.5mm]
& \multicolumn{4}{c}{TrafficBJ ($4 \rightarrow 4$)} & \multicolumn{5}{c}{Human3.6M ($4 \rightarrow 4$)}  \\
\cmidrule[0.25mm](lr){2-5}\cmidrule[0.25mm](lr){6-10}
\textbf{Method} &  \textbf{MSE$\pmb{\times}$100$\downarrow$} & \textbf{MAE}$\downarrow$ & \textbf{SSIM}$\uparrow$ & \textbf{FLOPs} & \textbf{MSE/10$\downarrow$} & \textbf{MAE/100$\downarrow$} & \textbf{SSIM}$\uparrow$  & \textbf{FVD}$\downarrow$ & \textbf{FLOPs}\\
\midrule[0.25mm]
ConvLSTM (2015)~\cite{shi2015convolutional} & 48.5 & 17.7 & 0.978 & 20.74G & 50.4 & 18.9 & 0.776 & 28.4 & 347.0G\\
PredRNN (2017)~\cite{wang2017predrnn} & 46.4 & 17.1 & 0.971 & 	42.40G & 48.4 & 18.9 & 0.781 & 24.7 & 704.0G\\
PredRNN++ (2018)~\cite{wang2018predrnn++} & 44.8 & 16.9 & 0.977 & 62.95G & $\times$ & $\times$ & $\times$ & $\times$ & 1,033G\\
E3D-LSTM (2019)~\cite{wang2019eidetic} & 43.2 & 16.9 & 0.979 & 98.19G	& 46.4 & 16.6 & 0.869 & 23.7 & 542.0G\\
MIM (2019)~\cite{wang2019memory} & 42.9 & 16.6 & 0.971 & 64.10G & 42.9 & 17.8 & 0.790 & 21.8 & 1,051G\\
PhyDNet (2020)~\cite{guen2020disentangling} & 41.9 & 16.2 & 0.982 & 5.60G & 36.9 & 16.2 & 0.901 & 18.3 & 19.1G\\
MotionRNN (2021)~\cite{wu2021motionrnn}& $\times$ & $\times$ & $\times$ & $\times$  & 34.2 & 14.8 & 0.846 & 18.3 & 49.5G\\
SimVP (2022)~\cite{gao2022simvp} & 41.4 & 16.2 & 0.982 & 3.61G & 31.6 & 15.1 & 0.904 & $\times$ & 197.0G\\
PredRNN-V2 (2023)~\cite{wang2022predrnn} & 45.6 & 16.8 & 0.980 & 42.63G	 & 36.3 & 17.7 & 0.863 & $\times$ & 708.0G\\
\midrule[0.25mm]
S-HR-VQVAE-disjoint (ours)   &  41.5   &  16.2 &  0.985   & \textbf{3.11G}  & 30.9 & 14.4 & 0.916 & 16.5 & \textbf{16.7G} \\
S-HR-VQVAE-joint (ours) & \textbf{40.3} & \textbf{15.2} & \textbf{0.993} & 4.07G& \textbf{30.4} & \textbf{12.4} & \textbf{0.939} & \textbf{15.2} & 17.4G\\
\bottomrule[0.5mm]
\multicolumn{7}{l}{($\uparrow$) means higher is better and ($\downarrow$) means lower is better.}
\end{tabular}
\end{table*}

\begin{table}
\scriptsize
\setlength{\tabcolsep}{10pt}
\centering
\caption {Results on Kitti dataset. S-HR-VQVAE with 3 layers was used with disjoint and joint training.}
\label{tbl:results_comparision_kitti}
\begin{tabular}{lccc}
\toprule[0.5mm]
& \multicolumn{3}{c}{Kitti ($4 \rightarrow 5$)} \\
\cmidrule[0.25mm](lr){2-4}
\textbf{Method} & \textbf{SSIM}$\uparrow$ & \textbf{LPIPS}$\downarrow$ & \textbf{FVD}$\downarrow$ \\
\midrule[0.25mm]
PredRNN (2017)~\cite{wang2017predrnn} & 0.475 & 0.629 & $\times$ \\
McNet (2018)~\cite{villegas2017decomposing} & 0.554 & 0.373 & $\times$ \\
NPVP (2023)~\cite{ye2023unified} & 0.661 & 0.279 & 134.69 \\
\midrule[0.25mm]
S-HR-VQVAE-disjoint (ours) & 0.673 & 0.188 & 127.04\\
S-HR-VQVAE-joint (ours) & \textbf{0.692} & \textbf{0.164} & \textbf{121.84} \\
\midrule[0.5mm]
& \multicolumn{3}{c}{Kitti ($5 \rightarrow 5$)} \\
\cmidrule[0.25mm](lr){2-4}
&  \textbf{SSIM}$\uparrow$ & \textbf{LPIPS}$\downarrow$ & \textbf{PSNR}$\uparrow$ \\
\midrule[0.25mm]
PhyDNet (2020)~\cite{guen2020disentangling} & 0.674 & 0.403 & 19.159 \\
LMC-Memory (2021)~\cite{lee2021video} & 0.660 & 0.410 & 18.692 \\
MotionRNN (2021)~\cite{wu2021motionrnn} & 0.652 & 0.384 & 18.931 \\
MIMO (2023)~\cite{ning2023mimo} & 0.703 & 0.308 & 19.616 \\
\midrule[0.25mm]
S-HR-VQVAE-disjoint (ours) & 0.845 & 0.187 & 19.774  \\
S-HR-VQVAE-joint (ours) & \textbf{0.861} & \textbf{0.114} & \textbf{21.877} \\
\bottomrule[0.5mm]
\multicolumn{4}{l}{($\uparrow$) means higher is better and ($\downarrow$) means lower is better.}
\end{tabular}
\end{table}

\subsection{Quantitative Analysis}
\label{subsec:quantitative}

In this study, we assess the performance of state-of-the-art video prediction methods on different datasets, providing a comprehensive overview of the advancements in the field.
In particular, Tables~\ref{tbl:results_comparision_KTH_MNIST}, ~\ref{tbl:results_comparision_traffic_human} and \ref{tbl:results_comparision_kitti} list state-of-the-art methods from 2015 to 2023 in a chronologically ascending order, highlighting thereby the evolution of the techniques over the years.
On the KTH Human Action task, PSNR and SSIM are reported by all competing techniques; whereas, LPIPS is provided for only a few methods.
On TrafficBJ and Human3.6M tasks, we report MSE, MAE, and SSIM as in~\cite{gao2022simvp, guen2020disentangling, wang2019memory}.
Finally, on the Kitti dataset, we report SSIM, LPIPS, FVD, and PSNR.
Here we report our results in order of complexity of the task (from KTH Human action to Kitti).

%%%% Table 3 analysis (KTH human action)
For the \textbf{KTH Human Action} task, from Table~\ref{tbl:results_comparision_KTH_MNIST}, it is evident that the proposed S-HR-VQVAE outperforms all methods, up to fRNN, in \emph{all} reported metrics.
Among methods introduced after fRNN, S-HR-VQVAE outperforms PredRNN++ on two metrics out of three, E3D-LSTM on all, Conv-TT-LSTM on all, PhyDNet on one out of two, Jin et al.~\cite{jin2020exploring} on two out of three, LMC-Memory on two out of three, V-3D-ConvLSTM across all, R-ST-ConvLSTM on one out of two, SimVP on one out of two, PredRNN-V2 on all, VPTR on two out of three, and NPVP on two out of three.
It can also be seen from Table~\ref{tbl:results_comparision_KTH_MNIST} that SimVP has the overall best PSNR, but our method outperforms it and achieves the best result in terms of SSIM.
%R-ST-ConvLSTM has the overall best SSIM on KTH Human Action but we outperform it on PSNR and LPIPS.
%S-HR-VQVAE outperforms V-3D-ConvLSTM on all reported metrics for KTH Human Action.
%SimVP achieves the best PSNR but not the best SSIM, which was obtained by Conv-TT-LSTM for KTH Human Action.
%Similarly, for LPIPS, VPTR has the best performance; however, the method is outperformed by SimVP and Conv-TT-LSTM in PSNR and SSIM, respectively. 
For LPIPS, NPVP has the best performance; however, the method is outperformed by our method both in terms of PSNR and SSIM, despite NPVP downsampling video frames to $64\times64$ rather than the typical $128\times128$.
It is noteworthy that S-HR-VQVAE achieves those results with a significantly lower number of parameters with respect to all other methods and ranks second in terms of computational efficiency (FLOPs). %compared to all other methods.

%%%% Table 4 analysis (TrafficBJ and Human3.6M)
On the \textbf{TrafficBJ} task, as detailed in Table~\ref{tbl:results_comparision_traffic_human}, S-HR-VQVAE exhibits exceptional performance, outperforming existing state-of-the-art methods on all evaluation metrics.
Our model particularly stands out by significantly outperforming methods such as PhyDNet and SimVP, achieving the highest scores across all metrics.
A similar trend is observed on the challenging task of \textbf{Human3.6M}, where S-HR-VQVAE again outperforms the current state-of-the-art approaches, leading in all evaluation metrics, especially in FVD (better temporal modeling) and FLOPs (computational efficiency).

%%%% Table 5 analysis (Kitti)
The performance of the S-HR-VQVAE on the challenging task of the \textbf{Kitti} dataset is detailed in Table~\ref{tbl:results_comparision_kitti}.
Unlike other tasks such as KTH Human Action, TrafficBJ, and Human3.6M, where the background is static, the Kitti dataset introduces a unique challenge with its dynamic and complex environments.
This complexity comes from the challenging driving scenes, where both the foreground and background are in motion.
This requires the prediction model to accurately handle multiple moving elements and rapidly changing landscapes, a significant shift from tasks where movement is mainly due to a single object against a constant background.

For the Kitti ($4 \rightarrow 5$) task, S-HR-VQVAE has demonstrated remarkable improvement over traditional models like PredRNN and McNet and more recent approaches such as NPVP.
It not only obtains better performance in SSIM, showing the best perceptual quality of predictions but also achieves the lowest LPIPS and a significantly better FVD, indicating superior performance in capturing both spatial and temporal aspects of the scenes.
Similarly, for the Kitti ($5 \rightarrow 5$) task, S-HR-VQVAE significantly outperforms other state-of-the-art models such as PhyDNet, LMC-Memory, MotionRNN, and MIMO.
It achieves higher SSIM and PSNR values, which indicates that it not only captures higher structural similarities between the predicted and actual frames but also maintains high-quality predictions across various frames.
The lower LPIPS also shows further evidence of S-HR-VQVAE's capability to preserve more accurate textural and detail-oriented features that are critical in dynamic scenes.

%%%% Conclusion of these two analysis
Referring to Tables~\ref{tbl:results_comparision_KTH_MNIST}, ~\ref{tbl:results_comparision_traffic_human} and \ref{tbl:results_comparision_kitti}, we can observe that the different metrics improve over the years.
Also, it can be argued that starting from 2018, all methods are quite competitive with one another, and it is not possible to indicate a single technique that performs the best on the video prediction task across all metrics.
Indeed, when we attempt to determine the best method, it can be seen that methods performing best in one metric are usually outperformed by other methods in other metrics, and therefore, it is essential to evaluate the results using all available metrics.
From this analysis, we can conclude that although some of the state-of-the-art methods outperform our method on a single metric, S-HR-VQVAE is more robust across all metrics for the considered tasks, especially for the challenging tasks of Human3.6M and Kitti, where S-HR-VQVAE outperforms the state-of-the-art methods across all metrics.
Ultimately, Tables~\ref{tbl:results_comparision_KTH_MNIST}, \ref{tbl:results_comparision_traffic_human}, and \ref{tbl:results_comparision_kitti} show the positive impact of jointly training HR-VQVAE and AST-PM across all datasets. While joint training introduces a slight increase in FLOPs compared to disjoint training, this marginal rise is outweighed by the significant benefits of joint training, particularly in improving spatiotemporal modeling, which contributes to the overall performance of the model.

The effectiveness of our approach can be further appreciated by considering the following qualitative analysis since objective metrics might not capture all aspects of the actual quality of the predicted sequences.

\subsection{Qualitative Analysis}

\begin{figure*}
\centering
\includegraphics[width=0.9\textwidth]{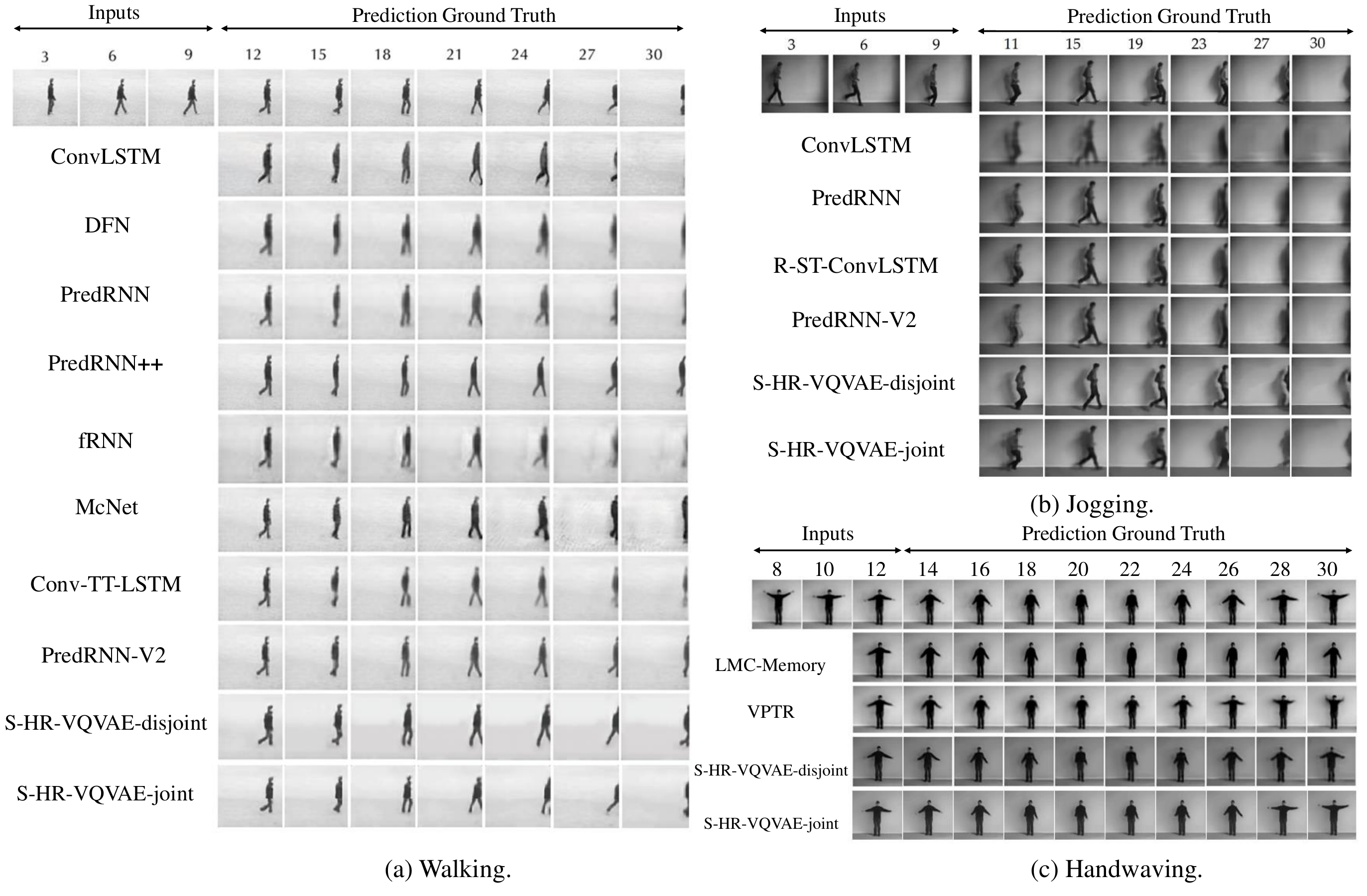}
\caption{Comparison of S-HR-VQVAE with state-of-the-art methods on KTH Human Moving Action dataset over three sequences (a, b, and c) that are commonly reported in the literature. It should be noted that 10 frames (1-10 in the figures) are given as input, and the next 20 frames (11-30 in the figures) are predicted.}
\label{fig:frame_by_frame_KTH}
\end{figure*}

\begin{figure}
\centering
\includegraphics[width=0.45\textwidth]{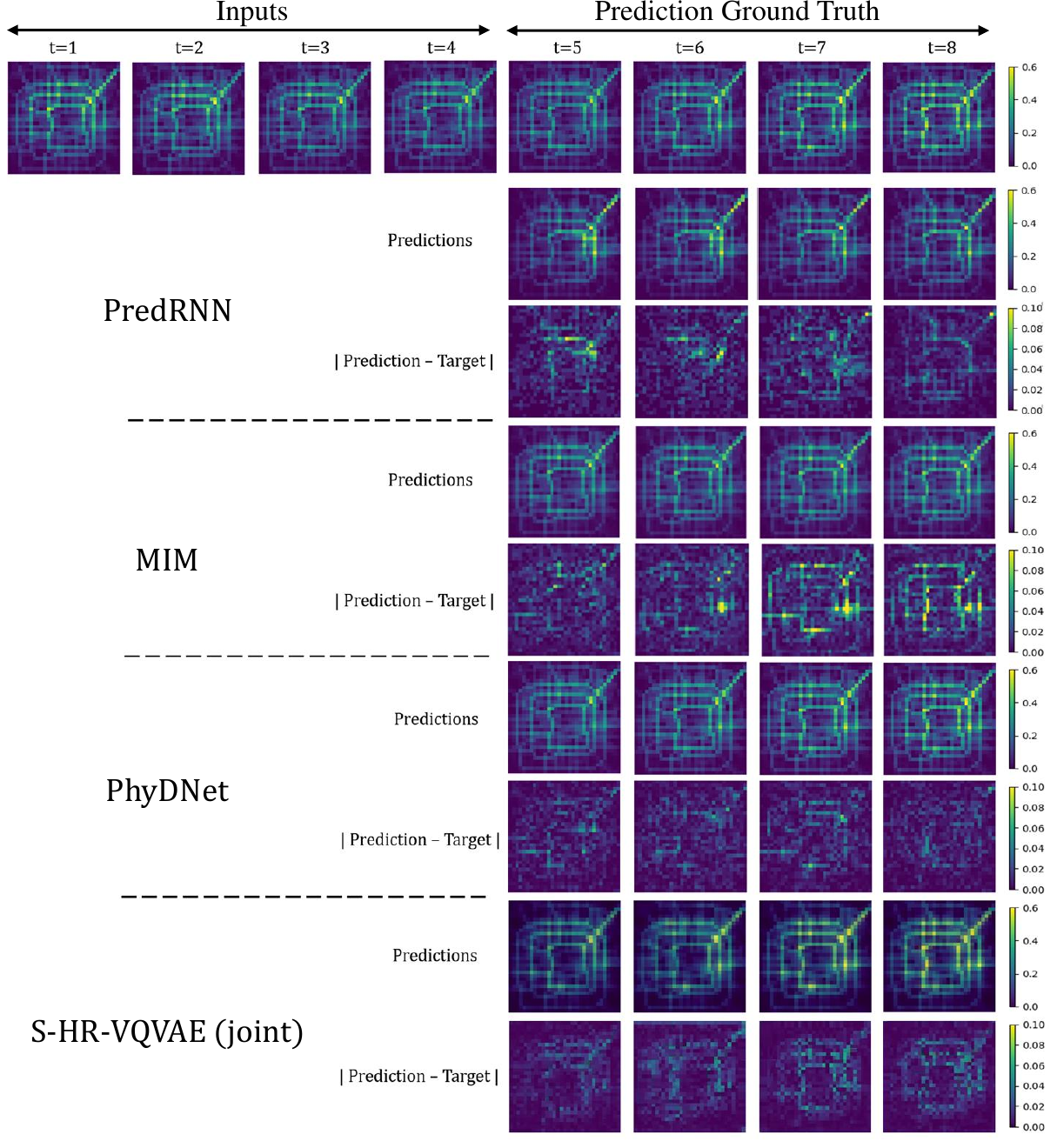}
\caption{Comparison of S-HR-VQVAE with state-of-the-art-methods on TrafficBJ dataset. It should be noted that 4 frames (1-4 in the figure) are given as input, and the next 4 frames (5-8 in the figure) are predicted.}
\label{fig:taxibj_frame_by_frame}
\end{figure}

\begin{figure}
\centering
\includegraphics[width=0.40\textwidth]{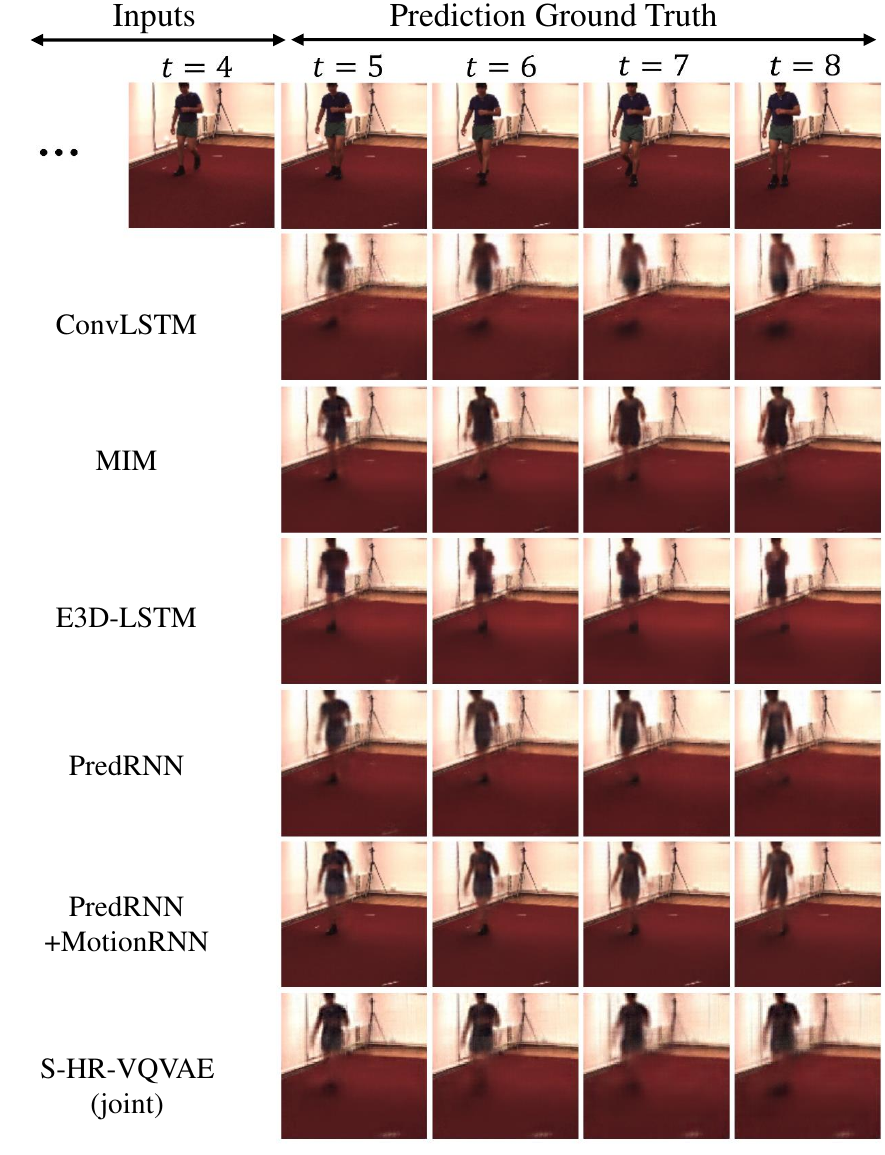}
\caption{Comparison of S-HR-VQVAE with state-of-the-art-methods on Human3.6M dataset. 4 frames (1-4 in the figure) are given as input, and the next 4 frames (5-8 in the figure) are predicted.}
\label{fig:human36_frame_by_frame}
\end{figure}

\begin{figure}
\centering
\includegraphics[width=0.48\textwidth]{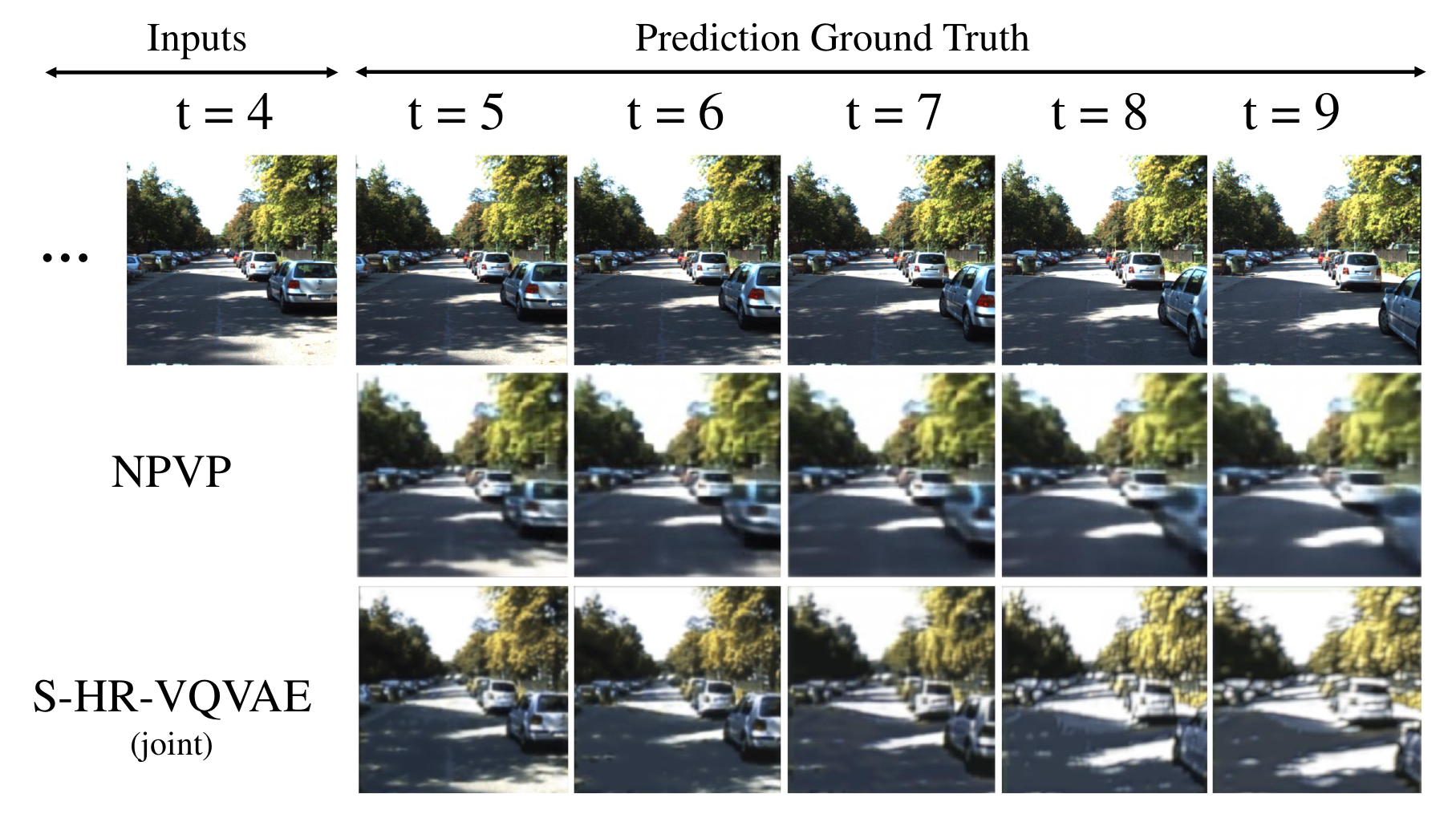}
\caption{Comparison of S-HR-VQVAE with the state-of-the-art method on the Kitti dataset, where 4 frames are given as input, and the next 5 frames are predicted.}
\label{fig:kitti_frame_by_frame}
\end{figure}

%%%% KTH Qualitative Analysis
Figure~\ref{fig:frame_by_frame_KTH} shows the predictions for different state-of-the-art methods and S-HR-VQVAE on the KTH Human Action dataset for three different activities: walking (panel a), jogging (panel b), and handwaving (panel c).
In the hand wave activity, for example,  hand movements are relatively fast, but S-HR-VQVAE can better predict the ground truth whilst avoiding blurry outputs, as shown in frames 28 and 30.
In the walking task, most methods do not predict well the position of the body and the legs, except for our method, PredRNN, PredRNN++, and PredRNN-V2 (see frames 27 and 30, for example).
However, our method produces sharper images and correctly predicts the location of both legs for these frames.
Finally, for the jogging task,  an overall better estimation of the location of the jogger is observed along with sharper images.

%The summarized qualitative analysis presented in Figure~\ref{fig:frame_by_frame_KTH} \hl{suggests that the joint training within our methodology leads to a notable enhancement in location prediction when compared to disjoint training. This improvement is evident across the majority of the frames. Nevertheless, it is important to acknowledge that while this improvement in location prediction is evident, it appears to be accompanied by a minor reduction in image sharpness in the reconstructed frames. This observation may provide insights into the relatively modest quantitative improvements observed in our results following the incorporation of joint training. Consequently, we extend our qualitative analysis to include the TrafficBJ, Human3.6M, and Kitti datasets, seeking to provide further insights into the implications of \textbf{joint training}}

%%%% TrafficBJ Qualitative Analysis
Figure~\ref{fig:taxibj_frame_by_frame} presents a qualitative analysis of the results obtained for TrafficBJ samples.
To enhance the clarity of our comparisons, we include visualizations of the differences between the predictions and the corresponding ground truth images.
S-HR-VQVAE demonstrates impressive performance in generating predicted frames when compared to the other models, as evidenced by the minimal intensity of differences observed.
It is noteworthy that S-HR-VQVAE obtains the best result on all metrics for this task.

%%%% Human3.6M Qualitative Analysis
The qualitative analysis presented in Figure~\ref{fig:human36_frame_by_frame} reveals that S-HR-VQVAE generates more precise predictions for motion positions and object sizes.
This observation underscores the efficacy of S-HR-VQVAE when applied to intricate real-world datasets.
S-HR-VQVAE better performance in predicting object positions and sizes can be attributed to the collaborative interaction between our spatiotemporal predictive model and the decoder, as stated in the objective function in Eq.~\ref{eq:joint_pixelCNN}.
%Such interaction significantly contributes to the prediction process accuracy.%, as well as the object's position and size.

%%%% Kitti Qualitative Analysis
The qualitative assessment on the Kitti dataset is depicted in Figure~\ref{fig:kitti_frame_by_frame}.
From the visual analysis, it is evident that S-HR-VQVAE exhibits finer details, such as intricate shadow patterns, leaf textures on trees, and more precise car features, compared to NPVP.
Moreover, S-HR-VQVAE significantly reduced blurry predictions compared to NPVP.
These observations align well with the quantitative findings presented in Table~\ref{tbl:results_comparision_kitti}, where S-HR-VQVAE outperforms NPVP across all evaluation metrics: SSIM, LPIPS, and FVD.
The higher SSIM score of S-HR-VQVAE indicates better structural similarity between predicted and ground truth frames, while the lower LPIPS value suggests reduced perceptual differences, underscoring the model's ability to generate more visually faithful predictions.
Furthermore, the significantly lower FVD score of S-HR-VQVAE compared to NPVP highlights its superiority in capturing temporal consistencies and minimizing artifacts.

%%%%%%%%% Conclusion of the analysis
We can summarise the outcome of the qualitative analysis as follows:
Although quantitative analysis is useful for understanding whether a sequence prediction technique is viable or not, objective measures by themselves may not reveal the actual capability of a technique.
State-of-the-art methods exhibit a varying sequence prediction quality across tasks, as observed, for example, in Figure~\ref{fig:frame_by_frame_KTH} despite the good numerical results reported in Table~\ref{tbl:results_comparision_KTH_MNIST}, whereas S-HR-VQVAE performs consistently across tasks. 
Moreover, the figure suggests joint training within our methodology leads to a significant enhancement in location prediction. This improvement is evident across the majority of the frames. Nevertheless, it is important to acknowledge that while this improvement in location prediction is evident, it appears to be accompanied by a minor reduction in image sharpness in the reconstructed frames. This observation may provide insights into the relatively modest quantitative improvements observed in our results following the incorporation of joint training.

\section{Discussion}
\label{sec:discussion}

\begin{figure}
\centering
\includegraphics[width=0.7\columnwidth]{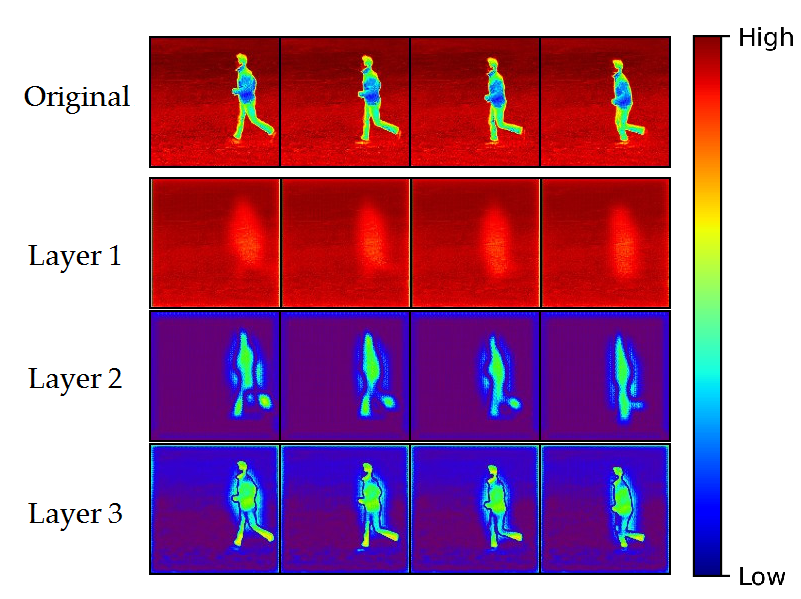}
\caption{Heatmap of reconstructions obtained from different layers of a 3-layer HR-VQVAE.}
\label{fig:heatmap}
\end{figure}

\subsection{Model Interpretability}
To facilitate the interpretation of latent representations produced by the model, we present heatmaps over various layers of HR-VQVAE in Fig.~\ref{fig:heatmap}.
Each heatmap highlights regions of significance within the reconstructed latent representation.
General information, i.e., background, is mainly captured in the first layer; the second layer focuses on the position of the foreground object, whereas the third layer is concerned with details of the moving objects.

\begin{figure}
\centering
\includegraphics[width=0.5\textwidth]{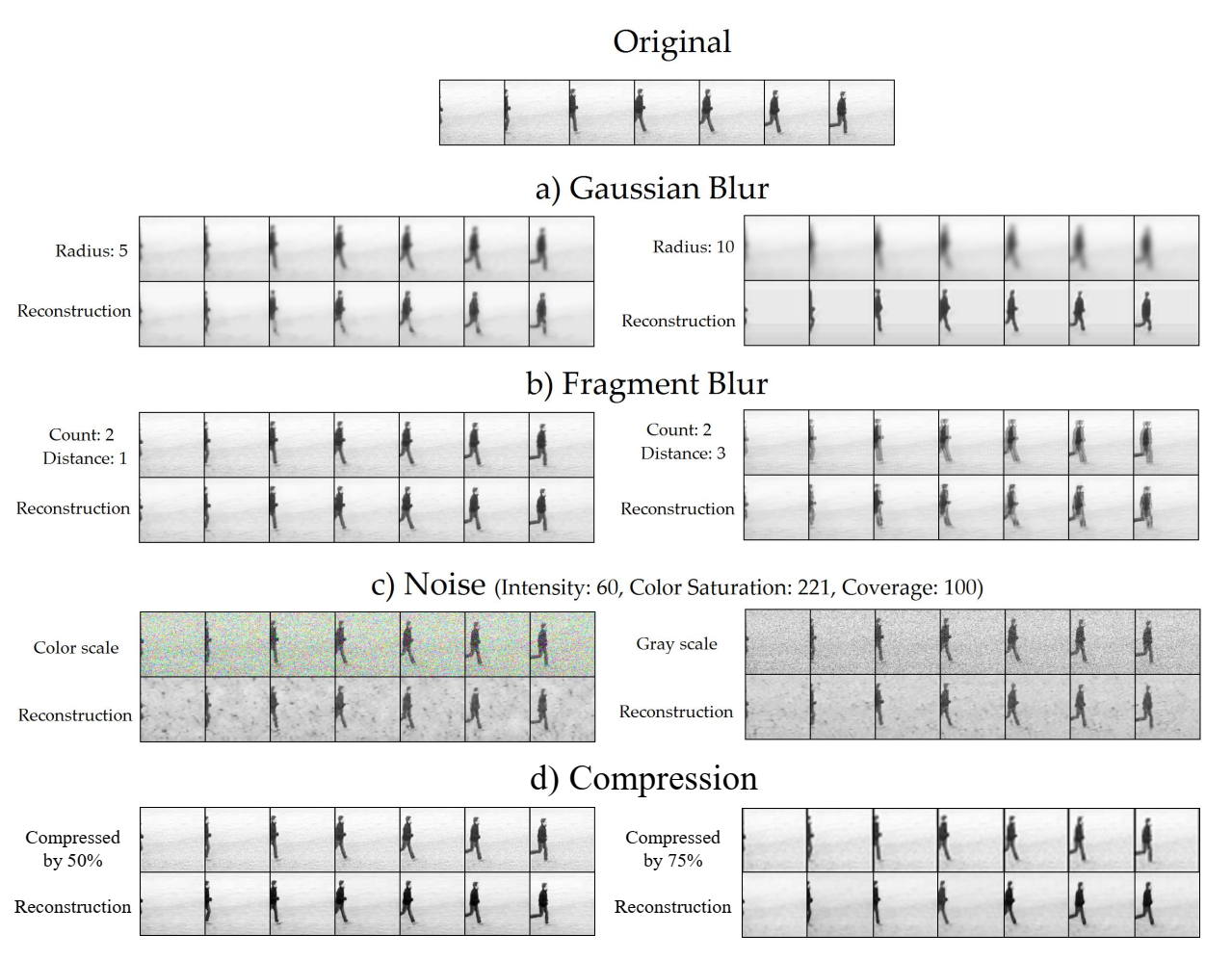}
\caption{Reconstructions by 3-layer HR-VQVAE. a) Gaussian Blur b) Fragment Blur c) Noise d) Compression. Zoom in to see more details.}
\label{fig:blur_fragment_noise}
\end{figure}

\subsection{Blur, Noise, and Compression}

To gain more insights into the effectiveness of S-HR-VQVAE against blurriness, we artificially corrupt some video sequences by injecting Gaussian Blur (Fig.~\ref{fig:blur_fragment_noise}-a) and Fragment Blur (Fig.~\ref{fig:blur_fragment_noise}-b).
The prediction results reported in those figures demonstrate that HR-VQVAE can successfully reduce blurriness while being able to reconstruct details in the images that were lost due to the blur effect.
In addition to blur mitigation, HR-VQVAE is also robust to noise, as shown in Fig.~\ref{fig:blur_fragment_noise}-c, where accurate sequence prediction is attained although the input frames were artificially corrupted with additive noise at different SNR levels.
Finally, we show the reconstruction of compressed images with two levels of compression ratio in Fig.~\ref{fig:blur_fragment_noise}-d, showcasing the HR-VQVAE's robustness against compression.
HR-VQVAE robustness against blur, noise, and compression in sequence prediction is especially valuable in applications where the quality of the predicted video frames is critical, such as autonomous driving.

\section{Conclusion}
\label{sec:conclusion}
In this study, we proposed a video prediction framework that combines the hierarchical vector quantization codebooks of the previously proposed HR-VQVAE with the novel autoregressive spatiotemporal predictive model (AST-PM).
We call this method sequential HR-VQVAE (S-HR-VQVAE).
We show how the proposed S-HR-VQVAE takes advantage of hierarchical frame modeling to model different levels of abstraction, enabling the system to capture both context and movements (details) in video frames with a fraction of the parameters used by competing models.
We show by extensive experimental evidence on the KTH Human Action, TrafficBJ, Human3.6M, and Kitti tasks that the model is very competitive with the state-of-the-art in video prediction, outperforming the best methods, at least in a subset of the available metrics (PSNR, SSIM, LPIPS, FVD, MSE, and MAE) with significantly lower number of parameters.
We  provide a detailed analysis of the properties of the model, including an analysis of its internal representations and its behavior concerning blurry and noisy input frames.
The proposed method is competitive for the video prediction task,  in terms of  performance, low complexity, and interpretability.

%\clearpage
%\newpage
\bibliographystyle{IEEEtran}
\bibliography{paper.bib}
\vspace{-1cm}
\begin{IEEEbiography}
[{\includegraphics[width=1in,height=1.25in,clip,keepaspectratio]{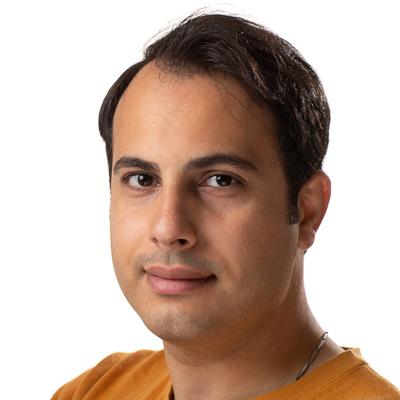}}]{Mohammad Adiban}
is a PhD candidate in Machine Learning at the Norwegian University of Science and Technology (NTNU). He holds a Bachelor's degree in Computer Engineering and a Master's degree in Artificial Intelligence from the Sharif University of Technology, awarded in 2017. In 2022, he conducted research as a visiting scholar at Monash University in Australia. Additionally, Mohammad is a co-founder of the company Connect Me and Senior Data Scientist at Bluware company. His research focuses on statistical machine learning, signal processing, computer vision, speech processing, biomedical applications, and cyber security.
\end{IEEEbiography}

\begin{IEEEbiography}
[{\includegraphics[width=1in,height=1.25in,clip,keepaspectratio]{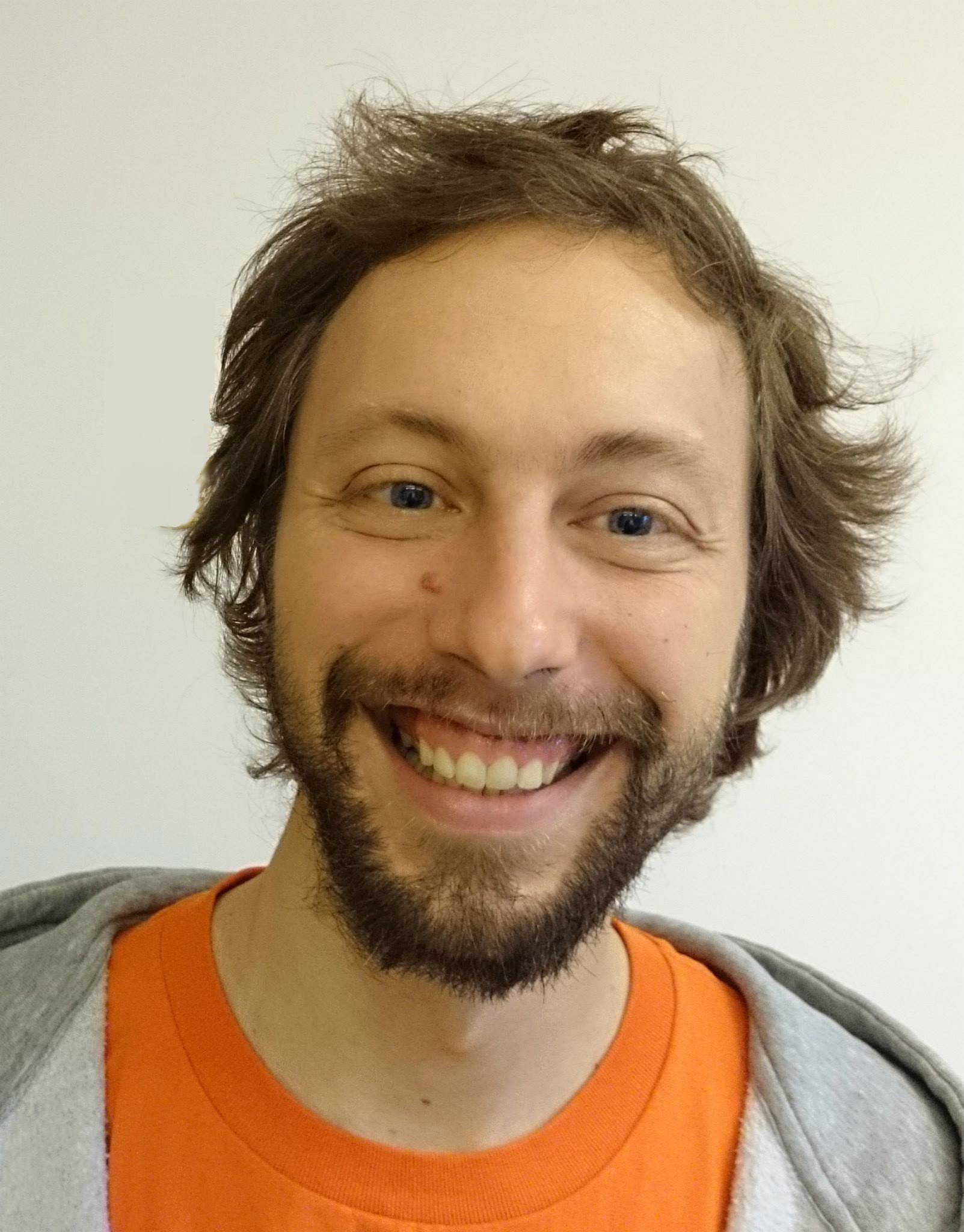}}]{Kalin Stefanov} is an ARC DECRA Fellow at the Faculty of Information Technology, Monash University, Melbourne, Australia. He received the MSc degree in Artificial Intelligence from the University of Amsterdam, Amsterdam, Netherlands and a PhD degree in Computer Science from KTH Royal Institute of Technology, Stockholm, Sweden. Prior to his current role, he was a Research Associate and Postdoctoral Research Scholar at the University of Southern California, Los Angeles, USA. His main research interests are machine learning, computer vision, and affective computing.
\end{IEEEbiography}

\begin{IEEEbiography}[{\includegraphics[width=1in,height=1.25in,clip,keepaspectratio]{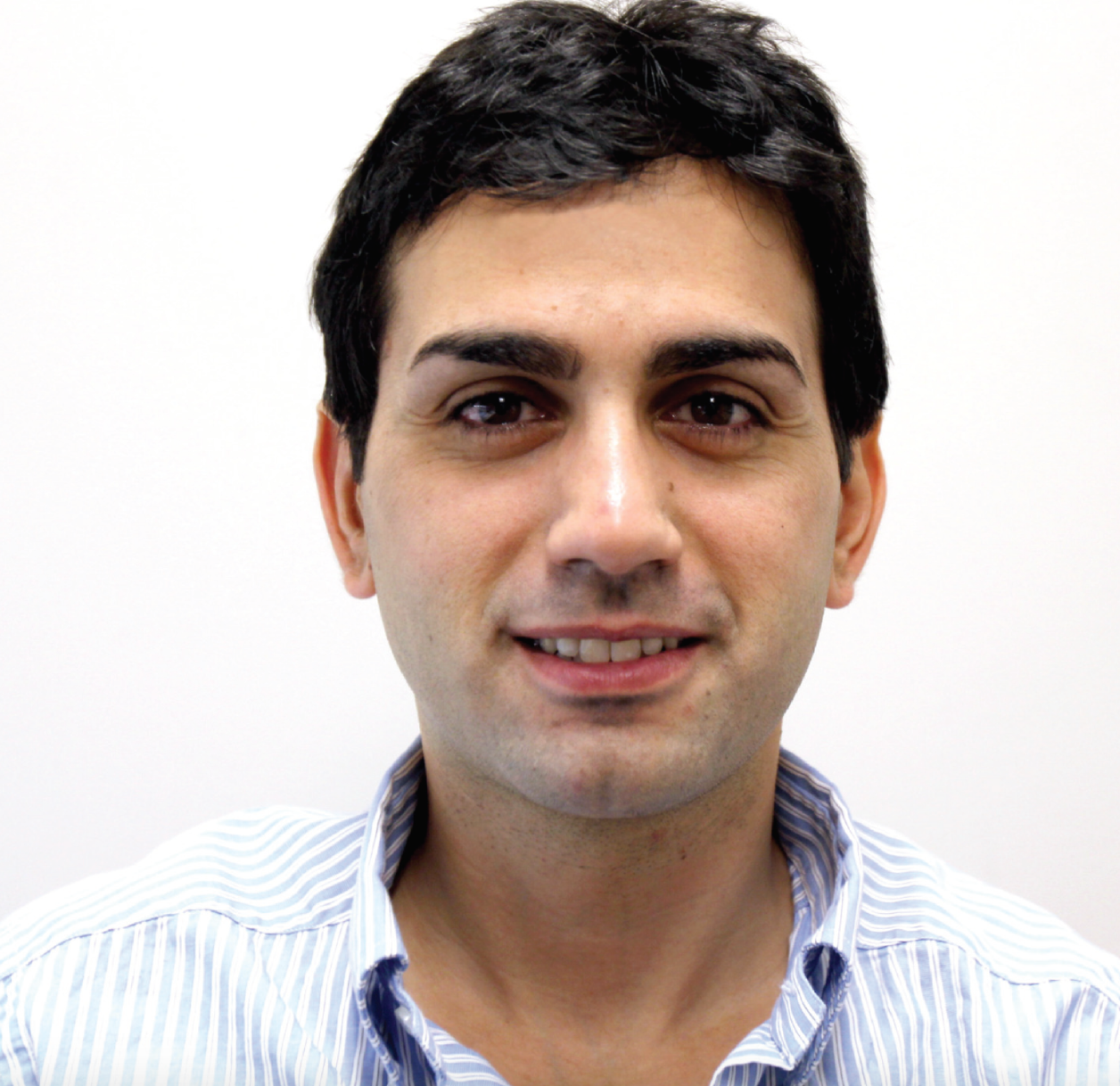}}]{Sabato Marco Siniscalchi}
(Senior Member, IEEE) is a FULL Professor with the University of Palermo,Palermo, Italy, an Adjunct Professor with the Norwegian University of Science and Technology (NTNU), and an Affiliate Faculty with the Georgia Institute of Technology. He received his doctorate degree in computer engineering from the University of Palermo, Palermo, Italy, in 2006. In 2006, he was a Postdoctoral Fellow with Ga Tech. From 2007 to 2010, he joined NTNU, Norway, as a Research Scientist. From 2010 to 2023, he was an Assistant Professor, first,  an Associate Professor, second, and a Full Professor, after, at Kore University. From 2017 to 2018, he was a Senior Speech Researcher with Siri Speech Group, Apple Inc., Cupertino CA, USA. He acted as an Associate Editor of the IEEE/ACM Transactions on Audio, Speech and Language Processing, from 2015 to 2019. Prof. Siniscalchi was an Elected Member of the IEEE SLT Committee from 2019 to 2022 and was re-elected in 2024.
\end{IEEEbiography}

\begin{IEEEbiography}
[{\includegraphics[width=1in,height=1.25in,clip,keepaspectratio]{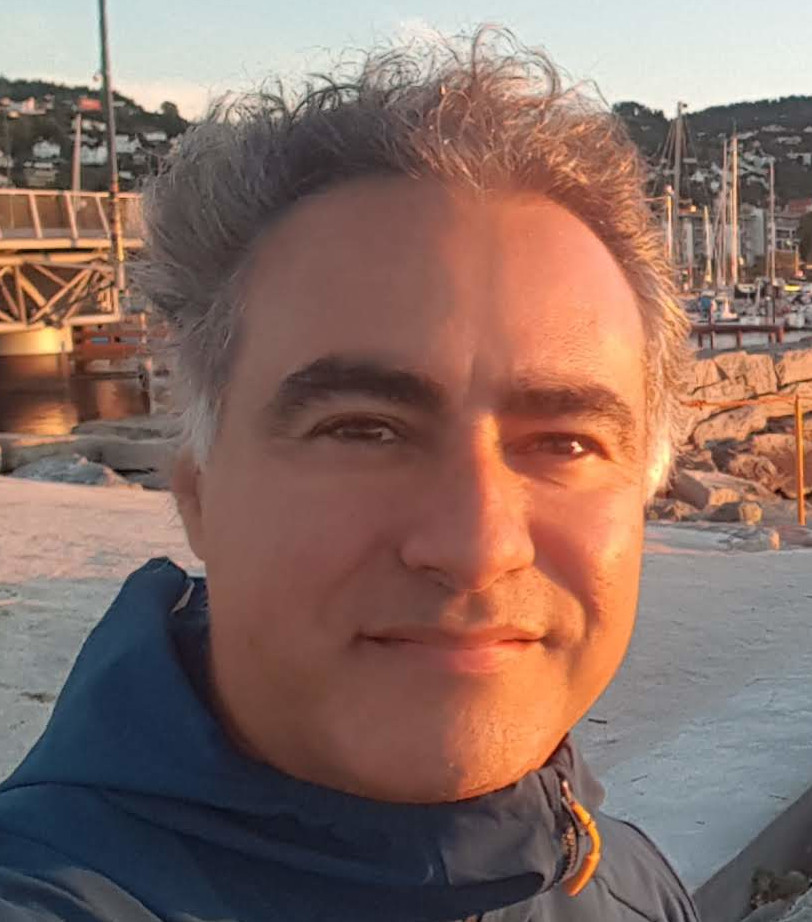}}]{Giampiero Salvi}
(Senior Member, IEEE) is a Full Professor at the Department of Electronic Systems at the Norwegian University of Science and Technology (NTNU), Trondheim, Norway, and Associate Professor at KTH Royal Institute of Technology, Department of Electrical Engineering and Computer Science, Stockholm, Sweden. Prof. Salvi received the MSc degree in Electronic Engineering from Università la Sapienza, Rome, Italy and the PhD degree in Computer Science from KTH. 
He was a post-doctoral fellow at the Institute of Systems and Robotics, Lisbon, Portugal.
He was a co-founder of the company SynFace AB, active between 2006 and 2016.
His main interests are machine learning, speech technology, and cognitive systems.
\end{IEEEbiography}

\end{document}